\documentclass{article}

 \usepackage[main, final]{neurips_2026}
\usepackage[utf8]{inputenc} % allow utf-8 input
\usepackage[T1]{fontenc}    % use 8-bit T1 fonts
\usepackage[colorlinks,
            linkcolor=red,
            anchorcolor=blue,
            citecolor=green
            ]{hyperref}
\usepackage{url}            % simple URL typesetting
\usepackage{booktabs}       % professional-quality tables
\usepackage{amsfonts}       % blackboard math symbols
\usepackage{nicefrac}       % compact symbols for 1/2, etc.
\usepackage{microtype}      % microtypography
\usepackage[dvipsnames]{xcolor}         % colors
\usepackage{colortbl}
\usepackage{xspace}
\usepackage{graphicx}
\usepackage{multirow}
\usepackage{makecell}
\usepackage{lscape}
\usepackage{bm}
\usepackage{algorithm}
\usepackage{algorithmic}
\usepackage{framed}
\usepackage{color}
\usepackage{bbm}
\usepackage{subfigure}
\usepackage{multirow} 
\usepackage{amssymb}
\usepackage{amsmath}
\usepackage{soul}
\usepackage{multirow}
\usepackage{makecell}
\usepackage{tabularx}
\usepackage{setspace}
\usepackage{utfsym}
\usepackage{pifont}
\usepackage{arydshln}
\usepackage{tcolorbox}
\usepackage{wrapfig}

\setcitestyle{square,comma,numbers}

\definecolor{lightgrayv}{HTML}{F4F3F8} % EEEDF3
\definecolor{grayv}{HTML}{707070}
\definecolor{redv}{HTML}{C00000}
\definecolor{bluev}{HTML}{0070C0}

\newcommand{\eg}{\emph{e.g.,}\xspace}
\newcommand{\ie}{\emph{i.e.,}\xspace}

\newcommand{\wrt}{{w.r.t.}\xspace}

\newcommand{\baby}{\textsc{EoupCT}\xspace}

\title{Estimating and Orthogonalizing Unknown Pre-training Gradients for Continual Fine-tuning of Large Language Models}

\author{
    Bing Wang$^{1,2}$, Changchun Li$^{1,2}$, Xin-Qiang Cai$^{3}$, Lin Yuanbo Wu$^{4}$, Ximing Li$^{1,2,3}$\thanks{Corresponding author}, \\ \textbf{Gang Niu}$^{3}$ \textbf{, Masashi Sugiyama}$^{3,5}$ \\
  $^{1}$ College of Computer Science and Technology, Jilin University \\
  $^{2}$ Key Laboratory of Symbolic Computation and Knowledge Engineering, MoE, Jilin University \\
  $^{3}$ RIKEN Center for Advanced Intelligence Project \quad
  $^{4}$ University of Warwick \\
  $^{5}$ Graduate School of Frontier Sciences, University of Tokyo \\
  \texttt{\{wangbing1416,changchunli93,liximing86\}@gmail.com}
}

\begin{document}

\maketitle

\begin{abstract}
  Continual fine-tuning is essential for large language models (LLMs) to dynamically adapt to real-world environments, yet it inevitably suffers from catastrophic forgetting, particularly the performance degradation of previous tasks and LLMs' general-purpose knowledge. Although existing methods, such as orthogonal gradient projection, mitigate the forgetting across various fine-tuning tasks, they fundamentally fail to preserve pre-training LLMs' inherent general-purpose knowledge because the original \textit{data and gradients of off-the-shelf pre-training LLMs required by these methods are strictly unknown and highly diverse}. To bridge this critical gap, we propose \baby, a novel framework designed to \textit{Estimate and Orthogonalize Unknown Pre-training gradients for Continual LLM fine-Tuning}. Specifically, \baby estimates pre-training gradients by dynamically generating pseudo data that is most susceptible to forgetting for new tasks through a learnable soft prompt equipped with Gumbel-Softmax relaxation. Furthermore, we formulate a multi-objective optimization problem and introduce a first-order efficient Pareto optimizer that jointly optimizes LLM parameters and the soft prompt, rigorously enforcing orthogonality between new task updates and the estimated pre-training gradients. Extensive experiments across multiple LLMs demonstrate that \baby effectively preserves both task-specific proficiency and inherent general-purpose knowledge, successfully mitigating the catastrophic forgetting.

\end{abstract}

\section{Introduction} \label{sec:1}

Recently, the community has witnessed the rapid development of \textit{large language models} (LLMs) in solving a variety of downstream tasks, \eg scientific Q\&A \citep{qwen2025qwen3} and mathematical reasoning \citep{guo2025deepseek,yan2026distribution}. However, in dynamic real-world environments, the ceaseless influx of knowledge across emerging tasks makes it difficult for off-the-shelf pre-training LLMs to timely adapt to new information. To bridge this gap, \textit{continual LLM fine-tuning} on a stream of new tasks presents a pivotal solution \citep{dou2023loramoe,wang2025continual}.

Despite the continuous acquisition of new task-specific knowledge, the primary challenge in continual LLM fine-tuning remains \textit{catastrophic forgetting}, that is, training on new tasks leads to a sustained performance decline on previously learned tasks and a degradation of the pre-training LLM's inherent general-purpose knowledge \citep{mccloskey1989catastrophic,luo2023an,shi2026continual,wang2026decomposing}.
To address this issue, a typical solution in the continual learning community is the \textit{orthogonal gradient optimization} approach \citep{wang2023orthogonal,wang2025continual,lu2025controlled}. Generally, this method involves storing the gradients of previously learned tasks and projecting the training gradients of the new task onto the orthogonal subspace spanned by these stored gradients, thereby mitigating interference between parameter updates for different tasks.
For instance, \textit{orthogonal low-rank adaptation} (OLoRA) \citep{wang2023orthogonal} allocates a distinct set of LoRA parameters \citep{hu2022lora} to each new task and enforces the orthogonality across the parameters of different task modules.

Although these orthogonalization methods are effective in mitigating catastrophic forgetting across various downstream tasks, \textit{they fail to account for the preservation of the pre-training LLMs' inherent general-purpose knowledge}, \eg commonsense knowledge and logic reasoning. In practice, maintaining pre-trained general-purpose knowledge is also computationally or architecturally infeasible for existing methods. Specifically, as \textit{the training data and gradients of off-the-shelf pre-training LLMs are typically unknown}, employing techniques such as gradient orthogonalization is rendered unrealistic. Furthermore, \textit{the tasks involved in pre-training LLM training are highly diverse} \citep{liao2025exploring}, even if recent techniques could estimate the underlying pre-training gradients by generating pseudo training data \citep{carlini2021extracting,kassem2024alpaca,nasr2025scalable}, identifying which gradients are most beneficial remains a critical challenge.

To address these challenges, we propose a novel method to \textit{Estimate and Orthogonalize Unknown Pre-training gradients for Continual LLM fine-Tuning}, namely \baby. Specifically, \baby involves two key steps. \textbf{First}, we estimate pre-training gradients by generating pseudo data \citep{carlini2021extracting,yu2023bag} through a learnable soft prompt \citep{lester2021the} utilized as the prefix to directly prompt the pre-training LLM. Since the direct auto-regressive generation of discrete tokens of the pseudo data is non-differentiable \wrt the soft prompt, we apply a Gumbel-Softmax relaxation \citep{jang2017categorical} to derive continuous token representations.
\textbf{Second}, we formulate a multi-objective optimization problem to jointly optimize the soft prompt and the task-specific LLM parameters.
Specifically, to optimize the LLM parameters, we not only perform continual fine-tuning on the new tasks but also enforce orthogonality between task-specific gradients, as well as between new task gradients and the estimated pre-training gradients. To optimize the soft prompt, we aim to generate pseudo data that is most susceptible to forgetting for each new task, while ensuring orthogonality among the estimated pre-training gradients to represent diverse pre-training tasks. To solve this complex multi-objective optimization problem, we further propose a first-order efficient Pareto optimizer, avoiding the computational burden of solving second-order optimization problems.

To evaluate the performance of \baby, we sample 15 natural language tasks from the \textit{SuperNI} dataset \citep{wang2022super}. We evaluate task-specific proficiency using the \textit{SuperNI} test subset and assess general-purpose knowledge via the \textit{MMLU} benchmark \citep{hendrycks2021measuring}. Generally, \baby achieves a 3.5\% improvement in average accuracy and a 3.7\% reduction in forgetting rate against SOTA results across 6 different pre-training LLMs, demonstrating that our approach consistently mitigates the catastrophic forgetting. \textit{Our source code is released in} \url{https://github.com/wangbing1416/EoupCT}.

In summary, our contributions can be summarized as the following three-folds:
\begin{itemize}
    \item We observe that while orthogonal gradient projection approaches mitigate interference between fine-tuning tasks, they struggle to preserve the pre-training LLMs' general-purpose knowledge, primarily due to the unknown and diverse nature of the pre-training gradients.
    \item To address unknown and diverse pre-training gradients during continual fine-tuning, we propose \baby, which estimates and orthogonalizes these pre-training gradients.
    \item Extensive experiments demonstrate that \baby significantly preserves both task-specific proficiency and general-purpose knowledge, thereby mitigating the catastrophic forgetting.
\end{itemize}

\section{Related Work}

In this section, we briefly summarize the related literature about recent continual LLM fine-tuning methods. 
Generally, existing research primarily mitigates the catastrophic forgetting issue during the continual LLM fine-tuning through several series of methods: \textit{orthogonal gradient projection}, which restricts parameter updates to lie within non-interfering parameter directions \citep{wang2023orthogonal,yang2025is,lu2025controlled,wang2025continual}.
\textit{Data rehearsal} involves continuously storing training samples from historical tasks and training them concurrently with the new task to mitigate their catastrophic forgetting \citep{sun2020lamol,wang2024inscl,huang2024mitigating,he2025dont}. For example, self-synthesized rehearsal (SSR) \citep{huang2024mitigating,wang2025self} utilizes a pre-training LLM to synthesize queries and employs a state-of-the-art LLM to generate corresponding responses as historical task data, which is then utilized for retraining during new task iterations. However, this approach remains unable to access pre-training data for the purpose of experience replay.
\textit{Parameter isolation}, \eg low-rank adaptation-based mixture-of-experts (LoRAMoE) \citep{dou2023loramoe,ma2024modula,zhao2024sapt,yang2026slora}, which allocates independent parameters for each task to isolate the influence between them \citep{liang2025gated,leng2025towards,wang2025remember,qiao2026merge,zhang2026grow}. While this approach decouples pre-trained weights from new task parameters, it necessitates maintaining a number of historical LLMs for routing during inference. Furthermore, achieving a balance between general-purpose knowledge and task-specific expertise remains challenging; as a unified model, even with parameter isolation, the integration of new tasks inevitably induces forgetting of general-purpose knowledge.

Among these, our work focuses on orthogonal gradient projection, a promising direction that ensures optimization directions do not interfere with general-purpose knowledge in pre-training weights. However, existing methods face two critical limitations. \textbf{First}, many current approaches do not operate directly on the gradients; instead, they often apply constraints to parameter weights, \eg OLoRA \citep{wang2023orthogonal} or hidden feature subspaces, \eg InfLoRA \citep{liang2024inflora}, which may not optimally represent the dynamic optimization landscape. \textbf{Second}, and more importantly, existing methods only consider the gradients of known sequential tasks within the training stream \citep{wang2025continual,lu2025controlled,xiong2026oplora}, which typically ignore the pre-training gradients because the pre-training data is unknown. This neglect leads to the erosion of the LLM's general-purpose knowledge during continual adaptation. To address these issues, we propose \baby that estimates the missing pre-training gradients, thereby enabling a more comprehensive orthogonal projection that protects both general-purpose and task-specific knowledge.

\vspace{-3pt}
\section{Our Proposed \baby}
\vspace{-3pt}
In this section, we introduce our proposed method \baby that orthogonally optimize LLMs.

\noindent
\textbf{Task definition of continual LLM fine-tuning.}
Let $\mathcal{F}_{\boldsymbol{\theta}^0}$ denote a pre-training LLM, which is pre-trained on a set of datasets $\mathcal{T}_0$ encompassing general-purpose knowledge. The goal of continual fine-tuning is to adapt this LLM to a sequence of downstream tasks. 
Specifically, we consider a sequence of datasets $[\mathcal{T}_1, \mathcal{T}_2, \dots, \mathcal{T}_T]$ representing $T$ different tasks, where each task $\mathcal{T}_t = \{(\mathbf{x}_i^t, \mathbf{y}_i^t)\}_{i=1}^{|\mathcal{T}_t|}$ consists of input prompts $\mathbf{x}_i^t$ and target outputs $\mathbf{y}_i^t$. The LLM, parameterized by $\boldsymbol{\theta}$ and initialized as $\boldsymbol{\theta}^0$, is then fine-tuned across this sequence by minimizing the empirical loss.
A critical issue in this process is the mitigation of catastrophic forgetting. This necessitates the preservation of both the general-purpose knowledge inherent in $\mathcal{F}_{\boldsymbol{\theta}^0}$, represented by $\mathcal{T}_0$, and the task-specific knowledge acquired from preceding tasks $\mathcal{T}_{1:t-1}$ during the fine-tuning of the current task dataset $\mathcal{T}_t$.

\noindent
\textbf{Formulation of orthogonal gradient projection methods.}
To fundamentally mitigate the interference between the current task $\mathcal{T}_t$ and previous tasks $\mathcal{T}_{1:t-1}$, orthogonal gradient projection restricts parameter updates to the orthogonal subspace spanned by previous inputs. Let $\boldsymbol{\theta}$ denote the trainable LLM parameters. During the fine-tuning on $\mathcal{T}_t$, we first compute the standard empirical gradient $\mathbf{g}_t = \nabla_{\boldsymbol{\theta}} \mathcal{L}(\mathcal{T}_t)$. Instead of applying this gradient directly, we project it using an orthogonal projection matrix $\mathbf{H}_{\perp}(\mathcal{T}_{1:t-1})$, which projects $\mathbf{g}_t$ onto the null space of the gradients $\mathbf{g}_{1:t-1}$ from all previous data $\mathcal{T}_{1:t-1}$. The parameter update rule is thus formulated as $\boldsymbol{\theta} \leftarrow \boldsymbol{\theta} - \eta \mathbf{g}_t \mathbf{H}_{\perp}(\mathcal{T}_{1:t-1})$, where $\eta$ is the learning rate. 
By mathematically ensuring that the projected gradient yields zero activation for any historical input $\mathbf{x} \in \mathcal{T}_{1:t-1}$, this method theoretically guarantees that the new parameter updates act as a null operator on previously acquired knowledge, therefore preventing catastrophic forgetting.

\subsection{Method Overview} \label{subsec:overview}

Although the orthogonal gradient projection methods prevent interference between the gradients of the current task $\mathcal{T}_{t}$ and previous tasks $\mathcal{T}_{1:t-1}$, they fail to account for orthogonality relative to the gradients of the pre-training task $\mathcal{T}_0$, as the latter remain \textit{unknown}. Therefore, it is challenging to enforce orthogonality between the gradients of new tasks and those of pre-training tasks, which leads to a degradation of general-purpose knowledge. To address this, the basic idea of \baby is to \textit{estimate pre-training gradients from the pre-training LLM $\mathcal{F}_{\boldsymbol{\theta}^0}$, and orthogonally optimize the gradients of tasks $\mathcal{T}_{1:T}$ against the pre-training gradients}.
Specifically, \baby is an end-to-end training strategy that consists of two pivotal modules: \textbf{\textit{pre-training gradient estimation}}, which directly prompts the pre-training LLM to generate pseudo data and estimates pre-training gradients, and \textbf{\textit{Pareto orthogonal optimization}}, which formulates a multi-objective optimization problem to ensure that the gradients of new tasks remain orthogonal to the estimated pre-training gradients.

Formally, given an off-the-shelf pre-training LLM $\mathcal{F}_{\boldsymbol{\theta}^0}$, and a trainable LLM parameterized by $\boldsymbol{\theta}$, \textit{pre-training gradient estimation} introduces a continuous $L$-length soft prompt $\mathbf{P}_t$ for the current task $\mathcal{T}_t$ into the pre-training LLM $\mathcal{F}_{\boldsymbol{\theta}^0}$ to generate pseudo data of the general-purpose knowledge distribution $\mathcal{T}_\text{0}$. Since discrete auto-regressive generation is non-differentiable during orthogonal optimization, we employ a Gumbel-Softmax relaxation to generate a continuous, fully differentiable surrogate sequence $\mathbf{S}(\mathbf{P}_t)$. This continuous generation maintains an unbroken computational graph, which is mathematically essential for the subsequent backpropagation optimization of the prompt $\mathbf{P}_t$.
Accordingly, we define a knowledge distillation loss $\mathcal{L}_\text{pre}(\mathbf{S}(\mathbf{P}_t))$ between the pre-training and trained LLMs on the pseudo data, and estimate the pre-training gradient $\mathbf{g}_\text{pre}(\mathbf{P}_t) = \nabla_{\boldsymbol{\theta}} \mathcal{L}_\text{pre}$.

Given the estimated gradients, \textit{Pareto orthogonal optimization} formulates the continual LLM fine-tuning process as an end-to-end multi-objective optimization problem. The objective is to jointly optimize the soft prompt $\mathbf{P}_t$ and the LLM parameters $\boldsymbol{\theta}$ to balance new task plasticity and general-purpose knowledge stability. 
Specifically, to achieve this goal, we present the following {\color{Blue} \textbf{\textit{C}}}\textit{onstraints}. Given a new task $\mathcal{T}_t$, 
{\color{Blue} \textbf{[C1] orthogonality}:} typically, addressing the forgetting requires ensuring gradient orthogonality not only between the new tasks
% \ie $\langle \nabla_{\Delta \mathbf{W}} \mathcal{L}_\text{new}(\mathcal{T}_t), \nabla_{\Delta \mathbf{W}} \mathcal{L}_\text{new}(\mathcal{T}_{1:t-1}) \rangle \rightarrow 0$, 
during the optimization of $\boldsymbol{\theta}$, but also between the new tasks and the estimated pre-training gradients.
% , \ie $\langle \nabla_{\Delta \mathbf{W}} \mathcal{L}_\text{new}(\mathcal{T}_t), \nabla_{\Delta \mathbf{W}} \mathcal{L}_\text{pre}(\mathbf{S}(\mathbf{P}_t)) \rangle \rightarrow 0$
Meanwhile, during the training of $\mathbf{P}_t$, orthogonality must be kept between the current and previous estimated pre-training gradients,
% \ie $\langle \nabla_{\Delta \mathbf{P}_t} \mathcal{L}_\text{pre}(\mathbf{S}(\mathbf{P}_t)), \nabla_{\Delta \mathbf{P}_t} \mathcal{L}_\text{pre}(\mathbf{S}(\mathbf{P}_{1:t-1})) \rangle \rightarrow 0$, 
to ensure that the pseudo data generated by different soft prompts represent distinct pre-training tasks.
{\color{Blue} \textbf{[C2] Vulnerability}:} we aim for the generated pseudo data sequence $\mathbf{S}(\mathbf{P}_t)$ to effectively expose the general-purpose knowledge that is most susceptible to being forgotten during the training of the current new task, \ie $\boldsymbol{\max}_{\mathbf{P}_t} \mathcal{L}_\text{new}(\mathcal{T}_t)$.
{\color{Blue} \textbf{[C3] Plasticity}:} given the generated pseudo data sequence $\mathbf{S}(\mathbf{P}_t)$ and new task data $\mathcal{T}_t$, we perform supervised fine-tuning on $\boldsymbol{\theta}$ using these datasets, \ie $\boldsymbol{\min}_{\boldsymbol{\theta}} \mathcal{L}_\text{new}(\mathcal{T}_t), \mathcal{L}_\text{pre}(\mathbf{S}(\mathbf{P}_t))$.

\begin{wrapfigure}{r}{0.56\textwidth}
\vspace{-12pt}
    \centering
    \includegraphics[width=0.56\textwidth]{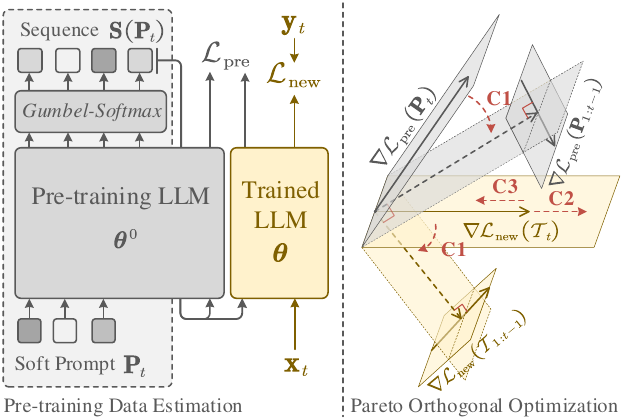}
    \vspace{-10pt}
    \caption{Framework of \baby with pre-training gradient estimation and Pareto orthogonal optimization.}
    \label{framework}
    \vspace{-10pt}
\end{wrapfigure}
Finally, our formulation in Appendix~\ref{subsec:hessian_to_first_order} demonstrates that the optimization of the aforementioned objective necessitates the computation of a second-order Hessian matrix, rendering it computationally intractable. Therefore, we propose a first-order Pareto optimizer to efficiently address this multi-objective optimization problem. For clarity, \textit{the overall framework of \baby is depicted in Fig.~\ref{framework}, and the complete algorithm of \baby is described in Appendix~\ref{sec.appendix.alg}.} In the following subsections, we provide a detailed introduction to pre-training gradient estimation, Pareto optimization objectives, and the efficient Pareto optimizer, respectively.

\vspace{-2pt}
\subsection{Pre-training Gradient Estimation}
\label{subsec:generative_replay}
\vspace{-2pt}
As motivated in Sec.~\ref{sec:1}, approximating the orthogonal complement of the general-purpose knowledge subspace requires access to the pre-training data distribution $\mathcal{T}_\text{0}$, which is unknown during continual LLM fine-tuning. To address this bottleneck, we propose a \textit{differentiable generation} mechanism that treats the off-the-shelf pre-training LLM $\mathcal{F}_{\boldsymbol{\theta}^0}$ itself as an implicit generative model of $\mathcal{T}_\text{0}$ \citep{carlini2021extracting,yu2023bag}.

Specifically, for the current fine-tuning task $\mathcal{T}_t$, we introduce a learnable $L$-length soft prompt $\mathbf{P}_t \in \mathbb{R}^{L \times D}$, where $D$ indicates the hidden size, that is prepended to $\mathcal{F}_{\boldsymbol{\theta}^0}$ to steer it toward generating sequences of the general-purpose knowledge distribution. However, standard auto-regressive decoding relies on discrete token selection, \eg top-$k$ sampling, which is non-differentiable and would sever the gradient flow $\nabla_{\mathbf{P}_t}$ essential for the optimization of the prompt $\mathbf{P}_t$.
To maintain a fully differentiable computational graph, we adopt the Gumbel-Softmax relaxation \citep{jang2017categorical} for differentiable decoding. Let $\mathbf{W}_\text{emb} \in \mathbb{R}^{|\mathcal{V}| \times D}$ denote the token embedding matrix, where $\mathcal{V}$ is the token vocabulary. We auto-regressively generate a sequence of $M$ soft token embeddings $\mathbf{S}(\mathbf{P}_t) = [\mathbf{e}_1, \mathbf{e}_2, \ldots, \mathbf{e}_M]$. 
At generation step $m \in \{1, \ldots, M\}$, the pre-training LLM $\mathcal{F}_{\boldsymbol{\theta}^0}$ processes the sequence $[\mathbf{P}_t, \mathbf{e}_1, \ldots, \mathbf{e}_{m-1}]$ to produce the next-token logits $\mathbf{z}_m \in \mathbb{R}^{|\mathcal{V}|}$. Instead of selecting a discrete token from $\mathbf{z}_m$, we compute a continuous probability vector via the Gumbel-Softmax relaxation as
\begin{equation}
    \pi_{m, i} = \frac{\exp \big( (z_{m, i} + \varepsilon_{m, i}) / \tau \big)}{\sum_{j=1}^{|\mathcal{V}|} \exp \big( (z_{m, j} + \varepsilon_{m, j}) / \tau \big)},
\end{equation}
where $\tau > 0$ is the temperature parameter (as $\tau \to 0$, $\boldsymbol{\pi}_m$ converges to a one-hot vector), and $\varepsilon_{m, i} = -\log(-\log(u_{m, i}))$ with $u_{m, i} \sim \text{Uniform}(0, 1)$ are Gumbel noise samples. The corresponding soft token embedding is obtained by computing the expected embedding under $\boldsymbol{\pi}_m$:
\begin{equation}
    \mathbf{e}_m = \boldsymbol{\pi}_m \mathbf{W}_\text{emb} = \sum \nolimits_{i=1}^{|\mathcal{V}|} \pi_{m, i} (\mathbf{W}_\text{emb})_i \in \mathbb{R}^D,
\end{equation}
which is appended to the prefix for the next decoding step. After $M$ iterations, this process yields the pseudo data $\mathbf{S}(\mathbf{P}_t)$. Accordingly, every operation from the initial soft prompt $\mathbf{P}_t$ to the final sequence $\mathbf{S}(\mathbf{P}_t)$ is strictly continuous and differentiable, making $\mathbf{P}_t$ optimizable. 
Accordingly, we define a knowledge distillation loss based on the \textit{mean squared error} between the latent representations of the pre-training and trained LLMs $\boldsymbol{\theta}^0$ and $\boldsymbol{\theta}$ on the pseudo data:
\begin{equation}
\label{eq:pre-training}
    \mathcal{L}_\text{pre}(\boldsymbol{\theta};\, \mathbf{S}(\mathbf{P}_t)) = \ell_\text{MSE} \left( \mathcal{F}_{\boldsymbol{\theta}^0}\big(\mathbf{S}(\mathbf{P}_t)\big),\; \mathcal{F}_{\boldsymbol{\theta}}\big(\mathbf{S}(\mathbf{P}_t)\big) \right),
\end{equation}
with the corresponding estimated pre-training gradient $\mathbf{g}_\text{pre}(\mathbf{P}_t) = \nabla_{\boldsymbol{\theta}} \mathcal{L}_\text{pre}$.

\vspace{-2pt}
\subsection{Multi-Objective Formulation of Pareto Orthogonal Optimization}
\label{subsec:pareto_moo}
\vspace{-2pt}

Given the pseudo data sequence $\mathbf{S}(\mathbf{P}_t)$ in Sec.~\ref{subsec:generative_replay} and the new task data $\mathcal{T}_t$, Pareto orthogonal optimization formalizes the three conditions {\color{Blue}\textbf{[C1-C3]}} outlined in Sec.~\ref{subsec:overview} into a unified \textit{multi-objective optimization} problem to jointly optimize the soft prompt $\mathbf{P}_t$ and the LLM parameter $\boldsymbol{\theta}$. Specifically, we formulate the following objectives.

\noindent
$\boldsymbol{\star}$ \textbf{Plasticity objective {\color{Blue} [C3]}.}
To learn the new task $\mathcal{T}_t$, we minimize the standard supervised fine-tuning loss over the task dataset. Formally, the plasticity objective is defined as the negative log-likelihood:
\begin{equation}
    \mathcal{L}_\text{new}(\boldsymbol{\theta}; \mathcal{T}_t) = -\frac{1}{|\mathcal{T}_t|} \sum \nolimits_{i=1}^{|\mathcal{T}_t|} \sum \nolimits_{k=1}^{|\mathbf{y}_i|} \log P_{\boldsymbol{\theta}}
    \left( y_{i,k}^t \mid \mathbf{x}_i^t, \mathbf{y}_{i, <k}^t \right),
    \quad \left( \mathbf{x}_i^t, \mathbf{y}_{i}^t \right) \in \mathcal{T}_t,
\end{equation}
where $P_{\boldsymbol{\theta}} \big( y_{i,k}^t \mid \mathbf{x}_i^t, \mathbf{y}_{i, <k}^t \big)$ denotes the probability of generating the $k$-th token $y_{i,k}^t$ conditioned on the input $\mathbf{x}_i^t$
. This yields the task-specific gradient $\mathbf{g}_\text{new} = \nabla_{\boldsymbol{\theta}} \mathcal{L}_\text{new}$.

\noindent
$\boldsymbol{\star}$ \textbf{Stability objective {\color{Blue} [C3]}.}
We minimize $\mathcal{L}_\text{pre}(\boldsymbol{\theta};\, \mathbf{S}(\mathbf{P}_t))$ in Eq.~\eqref{eq:pre-training} to keep the general-purpose knowledge stability.
To ensure that the pseudo data in Sec.~\ref{subsec:generative_replay} is informative and that the parameter update is orthogonal across previous gradients, we further impose the following three constraints.

\noindent
$\boldsymbol{\star}$ \textbf{Vulnerability {\color{Blue} [C2]}.}
The soft prompt $\mathbf{P}_t$ must dynamically expose general-purpose knowledge that is most susceptible to being forgotten during the current task $\mathcal{T}_t$. Instead of arbitrary estimation, we optimize $\mathbf{P}_t$ by maximizing $\mathcal{L}_\text{pre}$ on a \textit{virtually updated} $\boldsymbol{\theta}_\text{virt} = \boldsymbol{\theta} - \alpha \nabla_{\boldsymbol{\theta}} \mathcal{L}_\text{new}$. This virtual step simulates the destructive impact of the new task's gradient, forcing $\mathbf{P}_t$ to generate pseudo data that specifically exposes the general-purpose knowledge boundaries most damaged by the current task.

\noindent
$\boldsymbol{\star}$ \textbf{Pre-training \& new gradients orthogonality {\color{Blue} [C1]}.}
To resolve the forgetting, the new task gradient must not interfere with the currently estimated pre-training gradient $\mathbf{g}_\text{pre}(\mathbf{P}_t)$. This is inherently formulated as: finding an update direction that minimizes the new task loss without increasing the loss on the optimal $\mathbf{S}(\mathbf{P}_t^*)$, effectively enforcing orthogonality when gradient conflicts arise.

\noindent
$\boldsymbol{\star}$ \textbf{Pre-training \& pre-training gradients orthogonality {\color{Blue} [C1]}.}
To ensure diverse coverage of the general-purpose knowledge subspace, current pseudo data must not redundantly overlap with previous generations. We penalize the squared cosine similarity between the current protection gradient $\mathbf{g}_\text{pre}(\mathbf{P}_t)$ and historical pre-training gradients. This explicitly drives $\langle \mathbf{g}_\text{pre}(\mathbf{P}_t), \mathbf{g}_\text{pre}(\mathbf{P}_{1:t-1}) \rangle \to 0$, protecting distinct regions of the pre-training distribution.

\noindent
$\boldsymbol{\star}$ \textbf{New \& new gradients orthogonality {\color{Blue} [C1]}.}
To prevent forgetting of previous downstream tasks, the actual parameter update $\boldsymbol{\theta}_{t+1} - \boldsymbol{\theta}_t$ must reside entirely within the orthogonal null space of $\mathcal{M}_\text{new} = \{\mathbf{g}_\text{new}^1, \mathbf{g}_\text{new}^2, \ldots, \mathbf{g}_\text{new}^{t-1}\}$. This mathematically guarantees $\langle \boldsymbol{\theta}_{t+1} - \boldsymbol{\theta}_t, \mathbf{g}_\text{new}^i \rangle = 0$ for all historical gradients $\mathbf{g}_\text{new}^i \in \mathcal{M}_\text{new}$, ensuring absolute zero interference.

Integrating these objectives and constraints, the entire continual LLM fine-tuning process is cast as the following unified multi-objective optimization formulation:
\begin{align}
    \mathop{\boldsymbol{\min}}\limits_{\boldsymbol{\theta}} \ & \Big( \underbrace{\mathcal{L}_\text{new}(\boldsymbol{\theta};\, \mathcal{T}_t)}_{\text{New task plasticity}}, \;\; \underbrace{\mathcal{L}_\text{pre}\big(\boldsymbol{\theta};\, \mathbf{S}(\mathbf{P}_t^*)\big)}_{\text{General-purpose stability}} \Big) \label{eq:moo_objective} \\
    \text{s.t.} \ & \mathbf{P}_t^* = \boldsymbol{\arg} \mathop{\boldsymbol{\max}}\limits_{\mathbf{P}_t} \Big[ \underbrace{\mathcal{L}_\text{pre}\big(\boldsymbol{\theta}_\text{virt};\, \mathbf{S}(\mathbf{P}_t)\big)}_{\text{Vulnerability}} - \underbrace{\lambda \sum \nolimits_{i=1}^{t-1} \cos^2\!\big(\mathbf{g}_\text{pre}(\mathbf{P}_t),\, \mathbf{g}_\text{pre}(\mathbf{P}_i)\big)}_{\text{Pre-training \& pre-training gradients orthogonality}} \Big], \label{eq:constraint_adv} \\
    & \underbrace{\left\langle \boldsymbol{\theta}_{t+1} - \boldsymbol{\theta}_t,\; \nabla_{\boldsymbol{\theta}}\mathcal{L}_\text{pre}\big(\boldsymbol{\theta};\, \mathbf{S}(\mathbf{P}_t^*)\big) \right\rangle \le 0}_{\text{Pre-training \& new gradients orthogonality}} , \;
     \underbrace{\left\langle \boldsymbol{\theta}_{t+1} - \boldsymbol{\theta}_t,\; \mathbf{g}_\text{new}^{1:{t-1}} \right\rangle = 0}_{\text{New \& new gradients orthogonality}}. \label{eq:constraint_ortho}
\end{align}
This formulation encodes the goals of continual fine-tuning into a tractable mathematical structure. We next describe an efficient first-order optimizer for this second-order optimization problem.

\vspace{-3pt}
\subsection{Efficient Pareto Orthogonal Optimizer}
\label{subsec:efficient_solver}
\vspace{-3pt}
Directly solving the multi-objective optimization problem defined in Eqs.~\eqref{eq:moo_objective}, \eqref{eq:constraint_adv}, and \eqref{eq:constraint_ortho} is computationally intractable. This stems from the natural trade-off between vulnerability and stability; furthermore, explicitly coupling adversarial prompt optimization with new-task gradients would necessitate the computation of second-order Hessian matrices. Therefore, we resolve this by decoupling the optimizer into four efficient first-order steps, corresponding directly to our defined constraints. The complete algorithm of this efficient optimizer is shown in Appendix~\ref{sec.appendix.alg}.

\noindent
\textbf{Step 1:}
To satisfy both the vulnerability and the pre-training \& pre-training gradients orthogonality in Eq.~\eqref{eq:constraint_adv}, we optimize the soft prompt $\mathbf{P}_t$. A na\"{i}ve approach, maximizing $\mathcal{L}_\text{pre}$ at the current weights $\boldsymbol{\theta}_t$, only finds knowledge that the current model already struggles to preserve, completely ignoring the destructive effect of new tasks. The correct objective is to maximize the gradient conflict, \ie $\boldsymbol{\max}_{\mathbf{P}_t} \langle -\mathbf{g}_\text{new},\, \nabla_{\boldsymbol{\theta}} \mathcal{L}_\text{pre} \rangle$; however, computing its gradient \wrt $\mathbf{P}_t$ requires an expensive Hessian matrix.
We circumvent this via the \textit{virtual step approximation}: by a first-order Taylor expansion around the current weights $\boldsymbol{\theta}_t$, taking a virtual step $\boldsymbol{\theta}_\text{virt} = \boldsymbol{\theta}_t - \alpha\, \mathbf{g}_\text{new}$ changes the loss as
\begin{equation}
\label{eq:virtual_step}
    \mathcal{L}_\text{pre}(\boldsymbol{\theta}_\text{virt}) \approx \mathcal{L}_\text{pre}(\boldsymbol{\theta}_t) - \alpha \big\langle \mathbf{g}_\text{new},\, \nabla_{\boldsymbol{\theta}}\mathcal{L}_\text{pre}(\boldsymbol{\theta}_t) \big\rangle.
\end{equation}
Hence, maximizing $\mathcal{L}_\text{pre}(\boldsymbol{\theta}_\text{virt};\, \mathbf{S}(\mathbf{P}_t))$ over $\mathbf{P}_t$ is provably equivalent to maximizing the gradient conflict. Simultaneously, to enforce the orthogonality among estimated pre-training distributions, we subtract the historical independence penalty $\lambda \sum_{i=1}^{t-1} \cos^2\!\big(\mathbf{g}_\text{pre}(\mathbf{P}_t),\, \mathbf{g}_\text{pre}(\mathbf{P}_i)\big)$ from this objective during the backpropagation of $\mathbf{P}_t$. Once the optimal prompt $\mathbf{P}_t^*$ is obtained via this joint optimization, the true protection gradient is computed as $\mathbf{g}_\text{pre}^* = \nabla_{\boldsymbol{\theta}} \mathcal{L}_\text{pre}\big(\boldsymbol{\theta}_t;\, \mathbf{S}(\mathbf{P}_t^*)\big)$.

\noindent
\textbf{Step 2:}
With both $\mathbf{g}_\text{new}$ and $\mathbf{g}_\text{pre}^*$ in hand, we strictly enforce the orthogonality with new gradient in  Eq.~\eqref{eq:constraint_ortho} by constructing the null-space projection matrix of the historical new-task gradients $\mathcal{M}_\text{new}$:
\begin{equation}
\label{eq:null_space}
    \boldsymbol{\Pi}_\text{new} = \mathbf{I} - \mathbf{M}_\text{new} \big(\mathbf{M}_\text{new}^\top \mathbf{M}_\text{new}\big)^{-1} \mathbf{M}_\text{new}^\top,
\end{equation}
where $\mathbf{M}_\text{new}$ denotes the matrix by concatenating historical task gradients $\mathcal{M}_\text{new}$ as column vectors, and $\mathbf{I}$ is the identity matrix. Then, we project both raw gradients into this historically safe subspace:
\begin{equation}
\label{eq:orth_proj}
    \tilde{\mathbf{g}}_\text{new} = \boldsymbol{\Pi}_\text{new}\, \mathbf{g}_\text{new}, \quad 
    \tilde{\mathbf{g}}_\text{pre} = \boldsymbol{\Pi}_\text{new}\, \mathbf{g}_\text{pre}^*.
\end{equation}
This guarantees that any subsequent parameter update built from these projected gradients will have exactly zero inner product with any previously learned task.

\noindent
\textbf{Step 3:}
To satisfy the pre-training \& new gradients orthogonality in Eq.~\eqref{eq:constraint_ortho}, we must ensure the new task update does not increase the loss on the generated pseudo data. We compute the alignment score, if $\langle \tilde{\mathbf{g}}_\text{new},\, \tilde{\mathbf{g}}_\text{pre} \rangle \ge 0$, the two gradients are naturally aligned. If $\langle \tilde{\mathbf{g}}_\text{new},\, \tilde{\mathbf{g}}_\text{pre} \rangle < 0$, updating along $\tilde{\mathbf{g}}_\text{new}$ would damage the adversarially identified general knowledge. We resolve this by applying a Pareto-optimal projection that removes the conflicting component of $\tilde{\mathbf{g}}_\text{new}$ along the direction of $\tilde{\mathbf{g}}_\text{pre}$:
\begin{equation}
\label{eq:pareto_projection}
    \tilde{\mathbf{g}}_\text{new}^* = \tilde{\mathbf{g}}_\text{new} - \frac{\langle \tilde{\mathbf{g}}_\text{new},\, \tilde{\mathbf{g}}_\text{pre} \rangle}{\|\tilde{\mathbf{g}}_\text{pre}\|^2}\, \tilde{\mathbf{g}}_\text{pre}.
\end{equation}
A direct calculation confirms $\langle \tilde{\mathbf{g}}_\text{new}^*,\, \tilde{\mathbf{g}}_\text{pre} \rangle = 0$, mathematically guaranteeing that the modified task gradient avoids any interference with the currently estimated pre-training gradient.

\noindent
\textbf{Step 4:}
With the interference-free gradients established, the LLM parameters are updated as
\begin{equation}
\label{eq:final_update}
    \boldsymbol{\theta}_{\tau+1} = \boldsymbol{\theta}_\tau - \eta \big( \tilde{\mathbf{g}}_\text{new}^* + \tilde{\mathbf{g}}_\text{pre} \big),
\end{equation}
where $\eta > 0$ represents the learning rate. Because $\tilde{\mathbf{g}}_\text{new}^*$ and $\tilde{\mathbf{g}}_\text{pre}$ reside in the null space of $\mathcal{M}_\text{new}$ (by Step 2) and do not conflict with each other (by Step 3), this update direction $\boldsymbol{\theta}_{\tau+1} - \boldsymbol{\theta}_\tau$ simultaneously satisfies Eq.~\eqref{eq:constraint_ortho}, efficiently breaking the catastrophic forgetting.

\vspace{-4pt}
\section{Experimental Evaluation} \label{sec:experiments}
\vspace{-4pt}

In this section, we present the experimental performance of our \baby method.

\begin{table*}[t]
\centering
\renewcommand\arraystretch{1.0}
  \caption{Results of \baby under two task types: task-specific knowledge by \textit{SuperNI} and general-purpose knowledge by \textit{MMLU} (\textsc{Rou.} $\uparrow$, Acc. $\uparrow$, Fgt. $\downarrow$). The bold results represent the best scores.}
  \label{result}
  \small
  \setlength{\tabcolsep}{5pt}{
  \begin{tabular}{m{0.2cm}<{\centering}m{1.8cm}m{0.58cm}<{\centering}m{0.58cm}<{\centering}m{0.58cm}<{\centering}m{0.58cm}<{\centering}m{0.58cm}<{\centering}m{0.58cm}<{\centering}m{0.58cm}<{\centering}m{0.58cm}<{\centering}m{0.58cm}<{\centering}m{0.58cm}<{\centering}m{0.58cm}<{\centering}m{0.58cm}<{\centering}}
    \toprule
    & & \multicolumn{6}{c}{\textbf{\textit{SuperNI}}} & \multicolumn{6}{c}{\textbf{\textit{MMLU}} (average three task orders)} \\
    \cmidrule(r){3-8} \cmidrule(r){9-14}
    & \quad Method & \multicolumn{2}{c}{\textit{Order 1}} & \multicolumn{2}{c}{\textit{Order 2}} & \multicolumn{2}{c}{\textit{Order 3}} & \multicolumn{2}{c}{\textit{STEM}} & \multicolumn{2}{c}{\textit{Humanity}} & \multicolumn{2}{c}{\textit{Other}} \\
    \cmidrule(r){3-4} \cmidrule(r){5-6} \cmidrule(r){7-8} \cmidrule(r){9-10} \cmidrule(r){11-12} \cmidrule(r){13-14}
    & & \textsc{Rou.} & Fgt. &  \textsc{Rou.} & Fgt. & \textsc{Rou.} & Fgt. & Acc. & Fgt. & Acc. & Fgt. & Acc. & Fgt. \\
    \hline
    \multirow{6}{*}{\rotatebox{90}{\textbf{Qwen3-8B}}}
    & + LoRA & 50.1 & 7.7 & 48.2 & 9.8 & 49.9 & 8.5 & 62.6 & 3.5 & 71.3 & 7.6 & 72.5 & 4.4 \\
    & + LoRAMoE & 50.8 & 6.3 & 49.4 & 8.7 & 48.9 & 8.4 & 61.6 & 4.5 & 71.3 & 7.5 & 72.0 & 4.9 \\
    & + OLoRA & 51.2 & 6.7 & 47.4 & 9.4 & 50.1 & 7.6 & 63.3 & 2.9 & 72.0 & 6.9 & 72.9 & 4.0 \\
    & + GainLoRA & 50.0 & 6.6 & 49.7 & 8.0 & 49.2 & 8.1 & 62.7 & 3.4 & 72.7 & 6.2 & 72.4 & 4.4 \\
    & + CLoRA & 50.9 & 7.2 & 49.6 & 9.1 & 49.2 & 7.6 & 63.3 & 2.8 & 72.5 & 6.4 & 72.2 & 4.7 \\
    & \cellcolor{lightgrayv} + \textbf{\baby} & \cellcolor{lightgrayv} \textbf{52.9} & \cellcolor{lightgrayv} \textbf{4.4} & \cellcolor{lightgrayv} \textbf{52.1} & \cellcolor{lightgrayv} \textbf{6.0} & \cellcolor{lightgrayv} \textbf{51.5} & \cellcolor{lightgrayv} \textbf{5.7} & \cellcolor{lightgrayv} \textbf{64.7} & \cellcolor{lightgrayv} \textbf{1.4} & \cellcolor{lightgrayv} \textbf{75.9} & \cellcolor{lightgrayv} \textbf{3.0} & \cellcolor{lightgrayv} \textbf{75.5} & \cellcolor{lightgrayv} \textbf{1.4} \\

    \hline
    
    \multirow{6}{*}{\rotatebox{90}{\textbf{Qwen3-4B}}}
    & + LoRA & 45.1 & 13.7 & 40.8 & 16.7 & 48.6 & 9.2 & 56.4 & 4.5 & 64.0 & 8.6 & 64.2 & 5.7 \\
    & + LoRAMoE & 48.6 & 8.8 & 43.2 & 13.9 & 48.1 & 8.8 & 57.4 & 3.6 & 65.1 & 7.4 & 65.7 & 4.1 \\
    & + OLoRA & 48.9 & 9.0 & 46.0 & 10.6 & 47.9 & 9.6 & 58.1 & 2.9 & 65.4 & 7.1 & 65.9 & 3.9 \\
    & + GainLoRA & 48.4 & 9.0 & 44.9 & 12.1 & 47.8 & 9.6 & 55.9 & 5.0 & 64.8 & 7.8 & 65.3 & 4.6 \\
    & + CLoRA & 47.8 & 9.9 & 45.4 & 11.3 & 47.2 & 10.0 & 57.7 & 3.3 & 64.1 & 8.4 & 64.3 & 5.6 \\
    & \cellcolor{lightgrayv} + \textbf{\baby} & \cellcolor{lightgrayv} \textbf{50.7} & \cellcolor{lightgrayv} \textbf{6.0} & \cellcolor{lightgrayv} \textbf{50.3} & \cellcolor{lightgrayv} \textbf{6.1} & \cellcolor{lightgrayv} \textbf{50.5} & \cellcolor{lightgrayv} \textbf{6.2} & \cellcolor{lightgrayv} \textbf{60.2} & \cellcolor{lightgrayv} \textbf{0.7} & \cellcolor{lightgrayv} \textbf{67.4} & \cellcolor{lightgrayv} \textbf{5.2} & \cellcolor{lightgrayv} \textbf{68.0} & \cellcolor{lightgrayv} \textbf{1.9} \\

    \hline
    \specialrule{0em}{0.5pt}{0.5pt}
    \hline
    
    \multirow{6}{*}{\rotatebox{90}{\textbf{Llama3-8B}}}
    & + LoRA & 35.5 & 22.0 & 33.9 & 23.7 & 40.8 & 16.8 & 39.0 & 7.0 & 58.0 & 11.1 & 63.1 & 9.3 \\
    & + LoRAMoE & 42.9 & 14.7 & 36.7 & 20.5 & 37.5 & 20.2 & 40.3 & 5.7 & 59.1 & 10.0 & 64.3 & 8.1 \\
    & + OLoRA & 44.6 & 12.8 & 43.7 & 12.8 & 45.6 & 12.2 & 43.6 & 2.4 & 62.2 & 7.0 & 65.7 & 6.7 \\
    & + GainLoRA & 40.2 & 14.5 & 36.8 & 18.6 & 37.4 & 17.4 & 40.9 & 5.1 & 59.0 & 10.0 & 65.3 & 7.1 \\
    & + CLoRA & 41.6 & 14.7 & 37.2 & 18.2 & 41.5 & 13.5 & 42.1 & 3.9 & 62.0 & 7.0 & 64.9 & 7.5 \\
    & \cellcolor{lightgrayv} + \textbf{\baby} & \cellcolor{lightgrayv} \textbf{51.4} & \cellcolor{lightgrayv} \textbf{6.5} & \cellcolor{lightgrayv} \textbf{49.0} & \cellcolor{lightgrayv} \textbf{7.8} & \cellcolor{lightgrayv} \textbf{50.2} & \cellcolor{lightgrayv} \textbf{7.4} & \cellcolor{lightgrayv} \textbf{45.0} & \cellcolor{lightgrayv} \textbf{1.0} & \cellcolor{lightgrayv} \textbf{64.5} & \cellcolor{lightgrayv} \textbf{4.6} & \cellcolor{lightgrayv} \textbf{69.6} & \cellcolor{lightgrayv} \textbf{2.8} \\

    \hline
    
    \multirow{6}{*}{\rotatebox{90}{\textbf{Llama3-3B}}}
    & + LoRA & 37.6 & 17.7 & 39.3 & 14.4 & 43.7 & 10.7 & 38.9 & 1.7 & 56.0 & 6.1 & 59.5 & 4.2 \\
    & + LoRAMoE & 40.5 & 14.2 & 40.1 & 16.2 & 39.4 & 16.1 & 39.4 & 1.2 & 56.0 & 6.1 & 60.8 & 3.7 \\
    & + OLoRA & 41.6 & 13.5 & 36.4 & 20.0 & 36.7 & 18.3 & 39.2 & 1.4 & 55.6 & 6.5 & 60.0 & 2.9 \\
    & + GainLoRA & 38.0 & 16.3 & 38.3 & 18.1 & 43.0 & 11.6 & 38.4 & 2.2 & 55.8 & 6.3 & 59.3 & 4.3 \\
    & + CLoRA & 38.7 & 15.3 & 39.2 & 15.6 & 39.9 & 13.2 & 39.4 & 1.2 & 55.4 & 6.7 & 59.8 & 3.9 \\
    & \cellcolor{lightgrayv} + \textbf{\baby} & \cellcolor{lightgrayv} \textbf{44.0} & \cellcolor{lightgrayv} \textbf{10.5} & \cellcolor{lightgrayv} \textbf{45.1} & \cellcolor{lightgrayv} \textbf{9.2} & \cellcolor{lightgrayv} \textbf{46.5} & \cellcolor{lightgrayv} \textbf{8.2} & \cellcolor{lightgrayv} \textbf{41.4} & \cellcolor{lightgrayv} \textbf{-0.8} & \cellcolor{lightgrayv} \textbf{57.2} & \cellcolor{lightgrayv} \textbf{4.9} & \cellcolor{lightgrayv} \textbf{61.9} & \cellcolor{lightgrayv} \textbf{1.7} \\

    \hline
    \specialrule{0em}{0.5pt}{0.5pt}
    \hline
    
    \multirow{6}{*}{\rotatebox{90}{\textbf{Gemma2-9B}\ }}
    & + LoRA & 41.2 & 17.7 & 41.3 & 16.8 & 40.0 & 19.1 & 41.7 & 11.5 & 60.7 & 13.9 & 64.3 & 12.8 \\
    & + LoRAMoE & 48.8 & 10.2 & 36.6 & 23.0 & 45.5 & 13.7 & 46.8 & 7.4 & 65.7 & 8.9 & 68.7 & 8.4 \\
    & + OLoRA & 50.0 & 8.8 & 44.9 & 14.4 & 40.7 & 17.0 & 42.9 & 10.3 & 61.8 & 12.8 & 67.7 & 9.5 \\
    & + GainLoRA & 43.3 & 12.9 & 40.7 & 17.0 & 42.2 & 14.4 & 44.3 & 8.9 & 65.6 & 9.0 & 67.9 & 9.2 \\
    & + CLoRA & 42.8 & 13.3 & 40.8 & 15.5 & 42.5 & 12.7 & 45.1 & 8.1 & 64.8 & 9.8 & 64.8 & 12.3 \\
    & \cellcolor{lightgrayv} + \textbf{\baby} & \cellcolor{lightgrayv} \textbf{53.4} & \cellcolor{lightgrayv} \textbf{4.8} & \cellcolor{lightgrayv} \textbf{53.9} & \cellcolor{lightgrayv} \textbf{5.5} & \cellcolor{lightgrayv} \textbf{54.3} & \cellcolor{lightgrayv} \textbf{4.5} & \cellcolor{lightgrayv} \textbf{52.3} & \cellcolor{lightgrayv} \textbf{0.9} & \cellcolor{lightgrayv} \textbf{71.8} & \cellcolor{lightgrayv} \textbf{2.7} & \cellcolor{lightgrayv} \textbf{71.7} & \cellcolor{lightgrayv} \textbf{5.4} \\

    \hline
    
    \multirow{6}{*}{\rotatebox{90}{\textbf{Gemma2-2B}}}
    & + LoRA & 34.7 & 20.8 & 35.5 & 20.4 & 28.5 & 27.3 & 31.0 & 8.3 & 45.1 & 11.7 & 52.9 & 14.0 \\
    & + LoRAMoE & 36.0 & 20.9 & 34.4 & 22.2 & 38.3 & 17.2 & 33.0 & 6.3 & 48.9 & 7.9 & 59.6 & 7.3 \\
    & + OLoRA & 39.7 & 16.5 & 36.2 & 19.6 & 40.3 & 15.5 & 33.1 & 6.2 & 47.9 & 8.9 & 58.0 & 8.9 \\
    & + GainLoRA & 41.7 & 14.4 & 41.4 & 14.9 & 43.4 & 12.4 & 34.8 & 4.5 & 50.1 & 6.7 & 59.4 & 7.5 \\
    & + CLoRA & 37.5 & 17.7 & 39.2 & 16.8 & 38.8 & 17.1 & 34.9 & 4.4 & 47.4 & 9.3 & 58.1 & 8.8 \\
    & \cellcolor{lightgrayv} + \textbf{\baby} & \cellcolor{lightgrayv} \textbf{46.7} & \cellcolor{lightgrayv} \textbf{10.3} & \cellcolor{lightgrayv} \textbf{48.1} & \cellcolor{lightgrayv} \textbf{8.4} & \cellcolor{lightgrayv} \textbf{50.9} & \cellcolor{lightgrayv} \textbf{4.6} & \cellcolor{lightgrayv} \textbf{38.5} & \cellcolor{lightgrayv} \textbf{0.8} & \cellcolor{lightgrayv} \textbf{54.4} & \cellcolor{lightgrayv} \textbf{2.5} & \cellcolor{lightgrayv} \textbf{63.0} & \cellcolor{lightgrayv} \textbf{3.9} \\
    \bottomrule
    
  \end{tabular} }
  \vspace{-10pt}
\end{table*}

\vspace{-2pt} \noindent
\textbf{Experimental settings.}
To evaluate the performance of our method, we conduct experiments using \textit{SuperNI} \citep{wang2022super}, a large-scale multi-task instruction tuning dataset. Following \citet{zhao2024sapt}, we select 15 tasks organized into 3 different sequential orderings for continual LLM fine-tuning. Upon the completion of each task's fine-tuning phase, we evaluate the LLM on the corresponding \textit{SuperNI} test subsets to assess task-specific performance. Furthermore, we utilize 9 tasks across 3 categories from the \textit{MMLU} dataset \citep{hendrycks2021measuring} to benchmark the preservation of the LLM's general-purpose knowledge.
In our implementation, we utilize 6 pre-training LLMs across 3 distinct families, including Qwen3, Llama3, and Gemma2, employing \textit{low-rank adaptation} (LoRA) \citep{hu2022lora} with a rank of $r=4$ for parameter-efficient fine-tuning. We compare our approach against 5 baselines, encompassing a variety of continual fine-tuning strategies. Detailed dataset statistics, descriptions of the LLMs and baseline methods, implementation specifics, and formal evaluation metrics are provided in Appendix~\ref{sec:appendix.setup}.

\vspace{-3pt}
\subsection{Main Results}
\vspace{-3pt}
Table~\ref{result} presents the overall performance of \baby compared with a diverse set of continual LLM fine-tuning baselines across multiple LLM families and scales. We conduct fine-tuning across three distinct task sequences in Table~\ref{taskorder} generated using 3 different random seeds, reporting the task-specific performance on \textit{SuperNI} for each individual ordering. Meanwhile, the general-purpose knowledge performance on \textit{MMLU} is reported as the average across these three task sequences.

Generally, we can observe that \baby consistently achieves superior performance in both task-specific learning and general-purpose knowledge preservation. On \textit{SuperNI}, our method yields the highest average \textsc{Rouge} scores under all task orders while significantly reducing forgetting. For instance, on Qwen3-4B, \baby improves the \textsc{Rouge} of OLoRA from 48.9 to 50.7 under Order 1, and simultaneously reduces the forgetting from 9.0 to 6.0. Similar improvements are consistently observed across different task sequences, indicating that the proposed method is robust to the order of incoming tasks.
More importantly, \baby demonstrates a substantial advantage in mitigating catastrophic forgetting. Across all model families, the forgetting values are consistently lower than those of competing methods. This verifies that explicitly enforcing orthogonality with estimated pre-training gradients effectively prevents destructive interference during continual updates.

Meanwhile, \baby also shows remarkable improvements in preserving general-purpose knowledge on \textit{MMLU}. Compared to existing methods, which often suffer from severe degradation in general-purpose knowledge, \baby achieves both higher accuracy and lower forgetting across all categories. For example, on Gemma2-9B, \baby improves the STEM accuracy of OLoRA from 42.9 to 52.3 while reducing forgetting from 10.3 to 0.9. Notably, these gains are consistently observed across different domains, including Humanity and Other, suggesting that the estimated pre-training gradients effectively capture diverse general knowledge.

\vspace{-5pt}
\subsection{Sensitivity Analysis} \label{sec:sensitivity}

\begin{wrapfigure}{r}{0.54\textwidth}
\vspace{-25pt}
    \centering
    \includegraphics[width=0.545\textwidth]{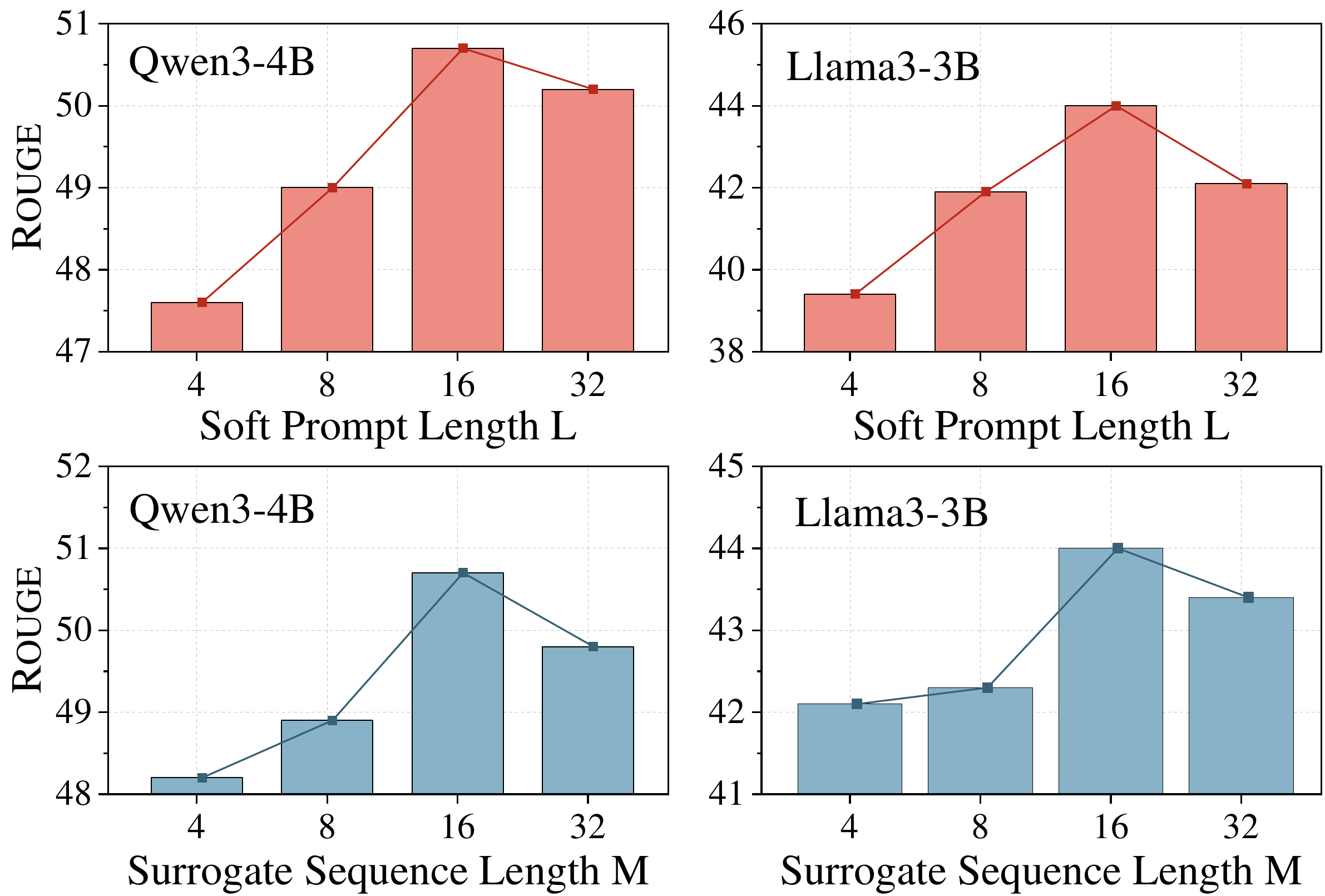}
    \vspace{-10pt}
    \caption{Sensitivity analysis of soft prompt length $L$ and surrogate sequence length $M$.}
    \label{sensitivity}
    \vspace{-10pt}
\end{wrapfigure}
We further investigate the sensitivity of \baby to two key hyper-parameters: the soft prompt length $L$ and the surrogate sequence length $M$. Specifically, we conduct experiments on Qwen3-4B and Llama3-3B, and report the performance in terms of \textsc{Rouge} on the \textit{SuperNI} benchmark under Order 1. Both $L$ and $M$ are selected from the set $\{4, 8, 16, 32\}$ to explore their impact on model performance.

The results show a consistent trend across both LLMs. When $L$ and $M$ are small, \eg 4 or 8, the model exhibits relatively weaker performance, suggesting that limited prompt capacity and short generated sequences are insufficient to adequately cover diverse pre-training knowledge. As $L$ and $M$ increase, the performance improves steadily, indicating that a richer soft prompt and longer surrogate sequences enable more expressive and representative reconstruction of the underlying pre-training distribution. However, when $L$ and $M$ become excessively large, \eg 32, the performance does not continue to improve and may even slightly degrade.

\subsection{Ablation Experiments}

We further conduct ablation studies to analyze the contribution of each component in \baby, as shown in Table~\ref{ablation}. Removing any component leads to a consistent degradation in both task-specific performance and general knowledge preservation, highlighting the necessity of the full design. In particular, removing the \textit{vulnerability objective, \ie w/o C2} results in a noticeable drop in \textsc{Rouge} and a sharp increase in forgetting. For example, on Qwen3-4B, the \textsc{Rouge} decreases from 50.7 to 47.6 while the forgetting increases from 6.0 to 10.3, indicating that identifying the most vulnerable pre-training knowledge is crucial for effective protection.

Similarly, \textit{removing the orthogonality constraints between pre-training gradients, \ie w/o P\&PO} leads to further performance degradation. On Llama3-3B, the \textsc{Rouge} drops from 44.0 to 42.1 and forgetting increases from 10.5 to 12.2, demonstrating that enforcing orthogonality with estimated pre-training gradients plays a key role in preserving general-purpose knowledge. When \textit{removing the orthogonality between new-task and pre-training gradients, \ie w/o N\&PO}, the LLM suffers from increased interference, resulting in higher forgetting across all settings. 
Moreover, \textit{removing the orthogonality among new-task gradients, \ie w/o N\&NO} leads to the most severe degradation, as it directly breaks the protection against inter-task interference. This confirms that maintaining orthogonality across sequential tasks is essential for stable continual learning. Across all ablations, we also observe consistent declines on \textit{MMLU} performance, indicating that each component contributes to preserving general-purpose knowledge.

\begin{table}[t]
\centering
\renewcommand{\arraystretch}{1.05}
\setlength{\tabcolsep}{4pt}
\caption{Ablation experiments of different constraints in \baby.}
\label{ablation}
\small
    \begin{tabular}{m{1.80cm}m{0.66cm}<{\centering}m{0.66cm}<{\centering}m{0.66cm}<{\centering}m{0.66cm}<{\centering}m{0.66cm}<{\centering}m{0.66cm}<{\centering}m{0.66cm}<{\centering}m{0.66cm}<{\centering}m{0.66cm}<{\centering}m{0.66cm}<{\centering}m{0.66cm}<{\centering}m{0.66cm}<{\centering}}
    \toprule
    \multirow{3}{*}{\quad Ablation} & \multicolumn{4}{c}{\textbf{Qwen3-4B}} & \multicolumn{4}{c}{\textbf{Llama3-3B}} & \multicolumn{4}{c}{\textbf{Gemma2-2B}} \\
    \cmidrule(r){2-5} \cmidrule(r){6-9} \cmidrule(r){10-13}
    & \multicolumn{2}{c}{\textit{SuperNI}} & \multicolumn{2}{c}{\textit{MMLU}} & \multicolumn{2}{c}{\textit{SuperNI}} & \multicolumn{2}{c}{\textit{MMLU}} & \multicolumn{2}{c}{\textit{SuperNI}} & \multicolumn{2}{c}{\textit{MMLU}} \\
    \cmidrule(r){2-3} \cmidrule(r){4-5} \cmidrule(r){6-7} \cmidrule(r){8-9} \cmidrule(r){10-11} \cmidrule(r){12-13}
    & \textsc{Rou.} & Fgt. & Acc. & Fgt. & \textsc{Rou.} & Fgt. & Acc. & Fgt. & \textsc{Rou.} & Fgt. & Acc. & Fgt. \\
    \midrule
    \rowcolor{lightgrayv} \textbf{\baby} & \textbf{50.7} & \textbf{6.0} & \textbf{60.2} & \textbf{0.7} & \textbf{44.0} & \textbf{10.5} & \textbf{41.4} & \textbf{-0.8} & \textbf{46.7} & \textbf{10.3} & \textbf{38.5} & \textbf{0.8} \\
    \quad w/o C2 & 47.6 & 10.3 & 59.0 & 2.0 & 42.3 & 13.0 & 40.4 & 0.2 & 44.8 & 10.1 & 36.9 & 2.3 \\
    \quad w/o P\&PO & 47.3 & 9.2 & 57.8 & 3.1 & 42.1 & 12.2 & 40.3 & 0.3 & 43.5 & 13.0 & 36.9 & 2.4 \\
    \quad w/o N\&PO & 47.4 & 8.7 & 57.9 & 3.0 & 41.9 & 12.2 & 40.0 & 0.6 & 44.2 & 13.3 & 36.6 & 2.6 \\
    \quad w/o N\&NO & 46.9 & 10.3 & 56.9 & 4.1 & 41.5 & 11.7 & 39.6 & 1.0 & 43.2 & 13.5 & 35.7 & 3.6 \\
    \bottomrule
    \end{tabular}
\vspace{-5pt}
\end{table}

% \vspace{-7pt}
\subsection{Forgetting Dynamics}

\begin{wrapfigure}{r}{0.55\textwidth}
\vspace{-20pt}
    \centering
    \includegraphics[width=0.555\textwidth]{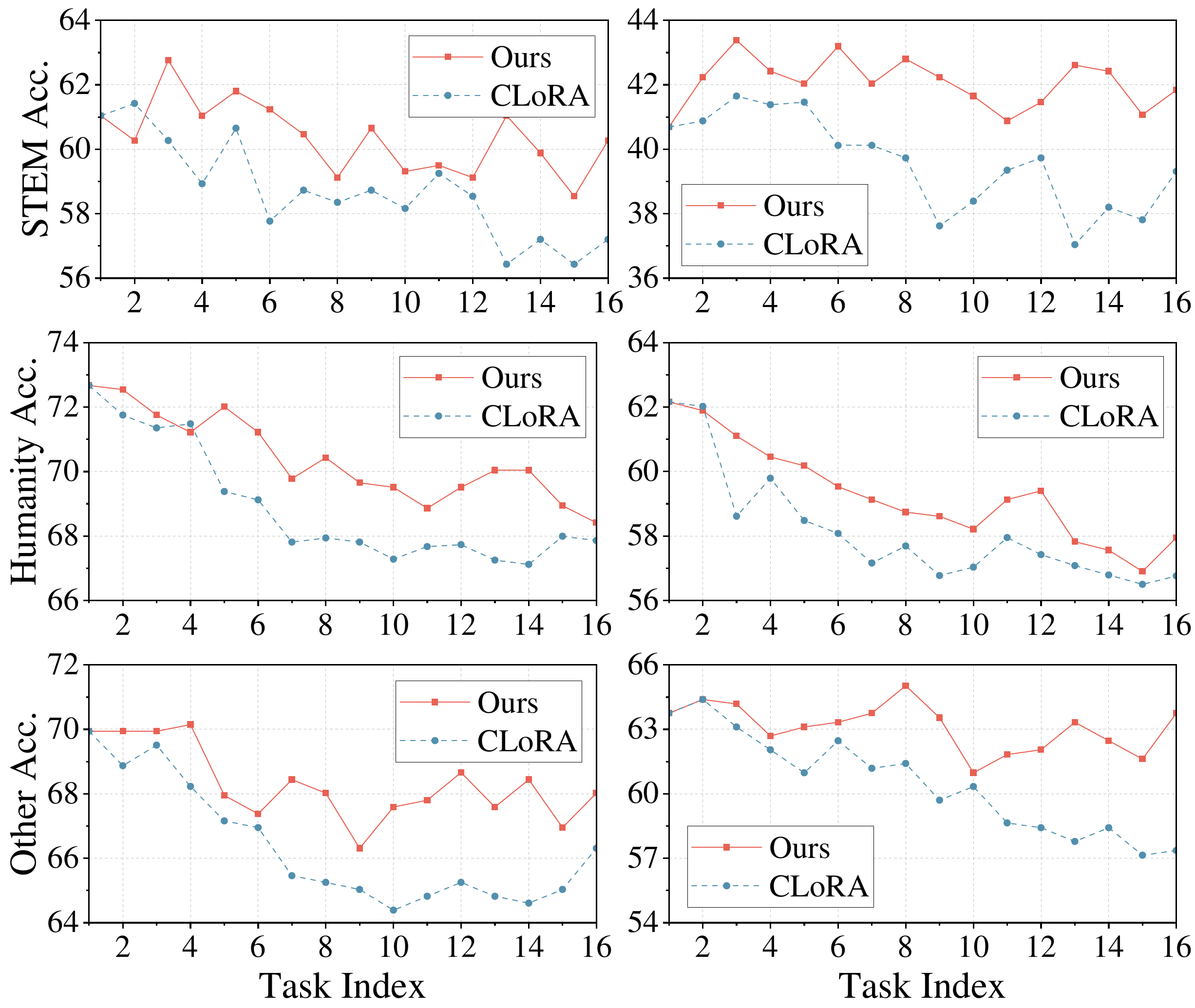}
    \vspace{-10pt}
    \caption{Forgetting dynamics of Qwen3-4B (left) and Llama3-3B (right) across \textit{MMLU}.}
    \label{evolution}
    \vspace{-9pt}
\end{wrapfigure}
To further analyze the forgetting behavior during continual fine-tuning, we track the evolution of general-purpose knowledge by evaluating the accuracy on three \textit{MMLU} categories as the \textit{SuperNI} tasks are sequentially learned. We compare \baby with the recent method CLoRA, focusing on how the performance changes throughout the training process.

The results reveal a clear advantage of \baby in terms of both the rate and extent of forgetting. As the number of learned tasks increases, CLoRA exhibits a steady decline in \textit{MMLU} accuracy, indicating progressive degradation of general-purpose knowledge. In contrast, \baby demonstrates a much slower decrease, maintaining consistently higher accuracy across all stages of training. This confirms that our method effectively mitigates the accumulation of forgetting over time.
More interestingly, we observe that \baby not only preserves knowledge but can even improve performance in certain stages, where the \textit{MMLU} accuracy temporarily surpasses that of the initial pre-trained LLM before any fine-tuning. This phenomenon suggests that the generated pseudo data and the associated orthogonal optimization not only prevent knowledge erosion but may also reinforce and refine useful representations within the LLM.

\vspace{-3pt}
\section{Conclusion}
In this paper, we address the challenge of catastrophic forgetting in the continual LLM fine-tuning, where existing methods fail to preserve inherent general-purpose knowledge due to the unknown and highly diverse nature of original pre-training data and gradients. To overcome this fundamental barrier, we propose \baby, a novel framework that dynamically estimates and orthogonalizes these unknown pre-training representations. Specifically, our approach leverages a learnable soft prompt equipped with a Gumbel-Softmax relaxation to generate the pseudo pre-training data most susceptible to forgetting. Furthermore, we formulate a multi-objective optimization problem solved by an efficient first-order Pareto optimizer, which jointly optimizes the soft prompt and task-specific LLM parameters to rigorously enforce gradient orthogonality without the burden of second-order computations. Extensive experiments across multiple LLMs demonstrate that \baby effectively maintains task-specific proficiency while preserving the models' broad general-purpose capabilities, successfully offering a robust solution to the catastrophic forgetting in continual LLM fine-tuning.

\section*{Acknowledgements}

This work was supported in part by the National Natural Science Foundation of China (No.62276113, 62372150) and JST ASPIRE Grant Number JPMJAP2405.

% \clearpage
\bibliography{custom}
\bibliographystyle{abbrvnat}
%%%%%%%%%%%%%%%%%%%%%%%%%%%%%%%%%%%%%%%%%%%%%%%%%%%%%%%%%%%%

\clearpage
\appendix

\section{Theoretical Proofs}
\label{sec:appendix_proofs}

In this section, we address the following three questions by providing theoretical proofs.

\newcommand{\hangingpar}[1]{%
  \noindent
  \hangafter=1
  \setlength{\hangindent}{2.5em}
  #1\par
}
\hangingpar{
\textcolor{Blue}{\textbf{RQ1.}} \textit{From the data perspective}, can continuous Gumbel-Softmax relaxation be utilized to estimate the pre-training distribution $\mathcal{T}_{0}$ composed of discrete token sequences?
}
\hangingpar{
\textcolor{Blue}{\textbf{RQ2.}} \textit{From the local optimization perspective}, can our proposed orthogonal projection method theoretically resolve the catastrophic forgetting?
}
\hangingpar{
\textcolor{Blue}{\textbf{RQ3.}} \textit{From the global generalization perspective}, can maintaining only a soft prompt and its generated surrogate sequences effectively preserve the vast repository of pre-trained knowledge?
}

\subsection{Can Gumbel-Softmax Relaxation Estimate Discrete Pre-training Sequences?}

In Sec.~\ref{subsec:generative_replay}, we utilize a continuous soft prompt $\mathbf{P}_t$ to generate a sequence of soft token embeddings $\mathbf{S}(\mathbf{P}_t)$. However, the authentic pre-training distribution $\mathcal{T}_0$ inherently resides in a discrete text space. We must mathematically guarantee that the stability objective $\mathcal{L}_\text{pre}$ evaluated on our differentiable soft sequence effectively bounds the expected loss of a true discrete text sequence, and that this approximation gap strictly vanishes as the Gumbel-Softmax temperature $\tau$ approaches zero.

\noindent \textbf{Theorem 1.} \textit{As formulated in Sec.~\ref{subsec:generative_replay}, let $\mathbf{W}_\text{emb} \in \mathbb{R}^{|\mathcal{V}| \times D}$ be the token embedding matrix. At decoding step $m$, let $\boldsymbol{\pi}_m(\tau) \in \mathbb{R}^{|\mathcal{V}|}$ be the continuous probability vector generated by the Gumbel-Softmax function at temperature $\tau$. Let $\mathbf{y}_m \in \{0,1\}^{|\mathcal{V}|}$ be the true discrete one-hot vector obtained by an ideal $\boldsymbol{\arg} \boldsymbol{\max}$ operation on the logits, e.g., greedy sampling. Assume the knowledge distillation loss function $\mathcal{L}_\text{pre}$ is $L$-Lipschitz continuous with respect to the input embeddings. The expected gap between the loss evaluated on the soft embedding and the discrete embedding is strictly bounded by $\mathcal{O}(\tau)$, i.e., a linear correlation.}

\textit{Proof.} We define the continuous soft token embedding as $\mathbf{e}_\text{soft} = \boldsymbol{\pi}_m(\tau) \mathbf{W}_\text{emb}$ and the corresponding true discrete token embedding as $\mathbf{e}_\text{discrete} = \mathbf{y}_m \mathbf{W}_\text{emb}$. By the definition of $L$-Lipschitz continuity \citep{khromov2024some}, the magnitude of the difference in the loss function is bounded by the Euclidean distance between the input vectors multiplied by the Lipschitz constant $L$, which gives 
\begin{equation}
    |\mathcal{L}_\text{pre}(\mathbf{e}_\text{soft}) - \mathcal{L}_\text{pre}(\mathbf{e}_\text{discrete})| \le L \|\mathbf{e}_\text{soft} - \mathbf{e}_\text{discrete}\|_2. \nonumber
\end{equation}
We expand this distance metric by substituting the definitions of the respective embeddings:
\begin{equation}
    \|\mathbf{e}_\text{soft} - \mathbf{e}_\text{discrete}\|_2 = \|(\boldsymbol{\pi}_m(\tau) - \mathbf{y}_m) \mathbf{W}_\text{emb}\|_2 \le \|\mathbf{W}_\text{emb}\|_{\max} \cdot \|\boldsymbol{\pi}_m(\tau) - \mathbf{y}_m\|_1, \nonumber
\end{equation}
where $\|\mathbf{W}_\text{emb}\|_{\max}$ denotes the maximum $L_2$-norm of any single token embedding vector within the vocabulary $\mathcal{V}$. A foundational property of the Gumbel-Softmax relaxation \citep{jang2017categorical} is that the expected $L_1$ distance between the soft probability distribution $\boldsymbol{\pi}_m(\tau)$ and the hard categorical one-hot vector $\mathbf{y}_m$ scales linearly with the temperature $\tau$ as $\tau$ becomes small. Consequently, we have $\mathbb{E} [ \|\boldsymbol{\pi}_m(\tau) - \mathbf{y}_m\|_1 ] \le C \cdot \tau$, where $C$ is a distribution-dependent constant. Taking the expectation on both sides of our Lipschitz inequality, we conclude:
\begin{equation}
    \mathbb{E} \big[ |\mathcal{L}_\text{pre}(\mathbf{e}_\text{soft}) - \mathcal{L}_\text{pre}(\mathbf{e}_\text{discrete})| \big] \le L \cdot \|\mathbf{W}_\text{emb}\|_{\max} \cdot C \cdot \tau = \mathcal{O}(\tau). \nonumber
\end{equation}
This mathematically proves that by annealing the temperature $\tau \to 0$ during the generative replay phase, our fully differentiable soft sequence seamlessly and accurately approximates the true discrete text distribution, introducing negligible approximation error while perfectly preserving the gradient flow for optimizing $\mathbf{P}_t$. \hfill $\square$

\subsection{Can \baby Theoretically Resolve the Catastrophic Forgetting?}

% While Sec.~\ref{sec2.3} provides a preliminary proof demonstrating that the orthogonal gradient projection method addresses the stability-plasticity dilemma formulated in Sec.~\ref{sec2.1}, this section offers a more comprehensive derivation. 
We show that our proposed \baby, which builds upon the principles of orthogonal gradient projection, effectively resolves the forgetting as well. 
In Step 3 of our Pareto orthogonal optimizer in Sec.~\ref{subsec:efficient_solver}, when the historically safe new task gradient $\tilde{\mathbf{g}}_\text{new}$ heavily conflicts with the protection gradient $\tilde{\mathbf{g}}_\text{pre}$, we project $\tilde{\mathbf{g}}_\text{new}$ onto the normal plane of $\tilde{\mathbf{g}}_\text{pre}$. We must guarantee that this specific projection is not a heuristic, but the exact geometric optimal solution that minimizes the new task loss while strictly prohibiting any increase in the general knowledge distillation loss.

\noindent \textbf{Theorem 2.} \textit{Let $\tilde{\mathbf{g}}_\text{new}$ and $\tilde{\mathbf{g}}_\text{pre}$ be the task and protection gradients as Eq.~\eqref{eq:orth_proj}, respectively, both already projected into the historical null space via $\boldsymbol{\Pi}_\text{new}$. If a gradient conflict is detected, i.e., $\langle \tilde{\mathbf{g}}_\text{new}, \tilde{\mathbf{g}}_\text{pre} \rangle < 0$, the proposed update direction $\tilde{\mathbf{g}}_\text{new}^* = \tilde{\mathbf{g}}_\text{new} - \frac{\langle \tilde{\mathbf{g}}_\text{new}, \tilde{\mathbf{g}}_\text{pre} \rangle}{\|\tilde{\mathbf{g}}_\text{pre}\|^2} \tilde{\mathbf{g}}_\text{pre}$ is the exact analytical solution to the constrained optimization problem: maximize the alignment with the new task gradient, subject to absolute zero interference with the estimated pre-training general-purpose knowledge.}

\textit{Proof.} We formulate the constrained gradient adjustment problem by seeking a modified update direction $\mathbf{d}$ that is geometrically as close as possible to the ideal new task gradient $\tilde{\mathbf{g}}_\text{new}$ (minimizing their Euclidean distance), subject to the hard constraint that moving along $\mathbf{d}$ will not increase the distillation loss $\mathcal{L}_\text{pre}$. To achieve this, $\mathbf{d}$ must be strictly orthogonal to the protection gradient $\tilde{\mathbf{g}}_\text{pre}$. Mathematically, this is expressed as $\mathbf{d}^* = \boldsymbol{\arg}\boldsymbol{\min}_{\mathbf{d}} (\|\mathbf{d} - \tilde{\mathbf{g}}_\text{new}\|^2 / 2)$ subject to $\langle \mathbf{d}, \tilde{\mathbf{g}}_\text{pre} \rangle = 0$. To solve this constrained objective, we construct the Lagrangian by introducing a Lagrange multiplier $\lambda \in \mathbb{R}$ for the equality constraint:
\begin{equation}
    \mathcal{L}(\mathbf{d}, \lambda) = \frac{1}{2} \|\mathbf{d} - \tilde{\mathbf{g}}_\text{new}\|^2 + \lambda \langle \mathbf{d}, \tilde{\mathbf{g}}_\text{pre} \rangle. \nonumber
\end{equation}
To locate the minimum, we compute the derivative of the Lagrangian $\mathcal{L}$ with respect to $\mathbf{d}$ and set it to the zero vector, yielding $\nabla_{\mathbf{d}} \mathcal{L}(\mathbf{d}, \lambda) = (\mathbf{d} - \tilde{\mathbf{g}}_\text{new}) + \lambda \tilde{\mathbf{g}}_\text{pre} = \mathbf{0}$, which rearranges to $\mathbf{d} = \tilde{\mathbf{g}}_\text{new} - \lambda \tilde{\mathbf{g}}_\text{pre}$. We then substitute this expression for $\mathbf{d}$ back into the strict hard orthogonality constraint $\langle \mathbf{d}, \tilde{\mathbf{g}}_\text{pre} \rangle = 0$:
\begin{equation}
    \langle \tilde{\mathbf{g}}_\text{new} - \lambda \tilde{\mathbf{g}}_\text{pre}, \tilde{\mathbf{g}}_\text{pre} \rangle = 0 \implies \langle \tilde{\mathbf{g}}_\text{new}, \tilde{\mathbf{g}}_\text{pre} \rangle - \lambda \|\tilde{\mathbf{g}}_\text{pre}\|^2 = 0. \nonumber
\end{equation}
Solving this algebraic equation for the Lagrange multiplier gives $\lambda = \frac{\langle \tilde{\mathbf{g}}_\text{new}, \tilde{\mathbf{g}}_\text{pre} \rangle}{\|\tilde{\mathbf{g}}_\text{pre}\|^2}$. Finally, substituting $\lambda$ back into the expression for $\mathbf{d}$ yields exactly our proposed projection update rule from Sec.~\ref{subsec:efficient_solver}:
\begin{equation}
    \mathbf{d}^* = \tilde{\mathbf{g}}_\text{new} - \frac{\langle \tilde{\mathbf{g}}_\text{new}, \tilde{\mathbf{g}}_\text{pre} \rangle}{\|\tilde{\mathbf{g}}_\text{pre}\|^2} \tilde{\mathbf{g}}_\text{pre} \equiv \tilde{\mathbf{g}}_\text{new}^*. \nonumber
\end{equation}
This derivation proves that our projection mechanism is the rigorous geometric Pareto optimum. It maximally preserves the plasticity required for learning $\mathcal{T}_t$ while mathematically enforcing absolute zero catastrophic forgetting of the estimated pre-training distribution. \hfill $\square$

\subsection{Can Maintaining Only a Soft Prompt Preserve Pre-trained Knowledge?}

\baby does not blindly replay the entire infinite pre-training distribution $\mathcal{T}_0$. Instead, we employ adversarial optimization in Eq.~\eqref{eq:constraint_adv} to locate a worst-case boundary sequence $\mathbf{S}(\mathbf{P}_t^*)$ and strictly protect it. A critical theoretical question emerges: why does protecting this specific, adversarially selected subset generated by only a soft prompt guarantee the protection of the vast, global pre-training distribution? We mathematically validate this using minimax generalization bounds.

\noindent \textbf{Theorem 3.} \textit{Let $\mathcal{T}_0$ denote the true global distribution of general-purpose knowledge. Let $\mathcal{P}$ define the expressive continuous space of the soft prompt $\mathbf{P}_t$. Assuming our adversarial solver successfully identifies the worst-case prompt $\mathbf{P}_t^* = \boldsymbol{\arg}\boldsymbol{\max}_{\mathbf{P}_t \in \mathcal{P}} \mathcal{L}_\text{pre}(\Delta\mathbf{W}_\text{virt}; \mathbf{S}(\mathbf{P}_t))$, and our Pareto-orthogonal update bounds this worst-case loss such that $\mathcal{L}_\text{pre}(\Delta\mathbf{W}_{t+1}; \mathbf{S}(\mathbf{P}_t^*)) \le \epsilon$, then the expected catastrophic forgetting over the entire true pre-training distribution is strictly bounded from above.}

\textit{Proof.} We formalize the total catastrophic forgetting across the pre-training LLM's general-purpose knowledge as the expected loss over the true pre-training distribution: $\text{Forgetting}_\text{global} = \mathbb{E}_{\mathbf{x} \sim \mathcal{T}_0} [ \mathcal{L}_\text{pre}(\Delta\mathbf{W}_{t+1}; \mathbf{x}) ]$. We operate under the standard assumption that the soft prompt space $\mathcal{P}$ is sufficiently expressive such that the generated continuous sequences can smoothly cover the support of the true distribution $\mathcal{T}_0$. Consequently, for any arbitrary actual data sample $\mathbf{x} \in \mathcal{T}_0$, there exists some prompt configuration $\mathbf{P}_{\mathbf{x}} \in \mathcal{P}$ capable of approximating it with an approximation error bounded by a small constant $\delta$, yielding the inequality $\mathcal{L}_\text{pre}(\mathbf{x}) \le \mathcal{L}_\text{pre}(\mathbf{S}(\mathbf{P}_{\mathbf{x}})) + \delta$. By the mathematical definition of the maximization operator in our vulnerability objective, the loss computed on any specific generated sequence $\mathbf{S}(\mathbf{P}_{\mathbf{x}})$ must be less than or equal to the loss evaluated on the adversarially optimized, worst-case sequence $\mathbf{S}(\mathbf{P}_t^*)$:
\begin{equation}
    \forall \mathbf{x} \in \mathcal{T}_0, \quad \mathcal{L}_\text{pre}(\mathbf{S}(\mathbf{P}_{\mathbf{x}})) \le \max_{\mathbf{P}_t \in \mathcal{P}} \mathcal{L}_\text{pre}(\mathbf{S}(\mathbf{P}_t)) = \mathcal{L}_\text{pre}(\mathbf{S}(\mathbf{P}_t^*)). \nonumber
\end{equation}
By chaining these two inequalities, we elegantly bound the loss of any genuine data point $\mathbf{x}$:
\begin{equation}
    \mathcal{L}_\text{pre}(\mathbf{x}) \le \mathcal{L}_\text{pre}(\mathbf{S}(\mathbf{P}_t^*)) + \delta. \nonumber
\end{equation}
Taking the expectation over $\mathbf{x} \sim \mathcal{T}_0$ strictly preserves this inequality, leading directly to $\mathbb{E}_{\mathbf{x} \sim \mathcal{T}_0} [ \mathcal{L}_\text{pre}(\mathbf{x}) ] \le \mathcal{L}_\text{pre}(\mathbf{S}(\mathbf{P}_t^*)) + \delta$. Because our Pareto-orthogonal solver (Sec.~\ref{subsec:efficient_solver}) explicitly enforces that the worst-case loss does not increase (\ie $\mathcal{L}_\text{pre}(\mathbf{S}(\mathbf{P}_t^*)) \le \epsilon$), the global expected forgetting is naturally bounded by:
\begin{equation}
    \text{Forgetting}_\text{global} \le \epsilon + \delta. \nonumber
\end{equation}
This mathematically proves that by explicitly hunting for and strictly protecting the vulnerability boundaries (the adversarial ``hard negatives''), \baby casts a robust, overarching protective net across the entire global distribution $\mathcal{T}_0$. \hfill $\square$

\textbf{Remark 3.1.} 
A potential critique of Theorem 3 is the assumption that the continuous prompt space $\mathcal{P}$ can densely cover the vast support of the true pre-training distribution $\mathcal{T}_0$. We clarify that our bound does not require global coverage. Instead, we rely on the concept of \textit{local adversarial representativeness}. For a given fine-tuning task at step $t$, the catastrophic forgetting is primarily driven by the parameter sub-space aligned with the task gradient. 
Therefore, optimizing $\mathbf{P}_t$ via the virtual step naturally forces the generated sequence $\mathbf{S}(\mathbf{P}_t)$ to act as an adversarial coreset. This coreset only needs to cover the specific local manifold of $\mathcal{T}_0$ that is most vulnerable to the current orthogonal update, bounded by a local approximation error $\delta_{local} \ll \delta_{global}$. As a result, the theoretical upper bound on forgetting remains tight and non-vacuous within the task-specific projection subspace.

\section{Complexity Analysis of \baby}
\label{sec:complexity_analysis}

In this section, we analyze the computational efficiency of the \baby framework. Appendix~\ref{subsec:hessian_to_first_order} mathematically demonstrates why directly solving the multi-objective optimization problem in Eqs.~\eqref{eq:moo_objective}, \eqref{eq:constraint_adv}, and \eqref{eq:constraint_ortho} inherently requires intractable second-order Hessian matrices, and how our efficient Pereto optimizer in Sec.~\ref{subsec:efficient_solver} reduces this to a first-order optimization; Appendix~\ref{subsec:time_complexity} provides a step-by-step time complexity comparison between continual fine-tuning baselines and \baby; Appendix~\ref{subsec:empirical_budget} empirically shows the computational budgets of \baby.

\subsection{Efficient First-Order Pareto Optimization} 
\label{subsec:hessian_to_first_order}

The core of our vulnerability objective {\color{Blue}\textbf{[C2]}} is to explicitly find a soft prompt $\mathbf{P}_t$ that generates data maximizing the gradient conflict with the new task. Mathematically, this means maximizing the negative inner product between the new task gradient $\mathbf{g}_\text{new}$ and the protection gradient $\nabla_{\boldsymbol{\theta}} \mathcal{L}_\text{pre}$. We now demonstrate why directly optimizing this requires a Hessian, and how we avoid it.

\noindent \textbf{Second-order optimization bottleneck.} 
Let us formalize the naive objective. To find the worst-case prompt $\mathbf{P}_t$, we want to solve:
\begin{equation}
    \mathop{\boldsymbol{\max}}\limits_{\mathbf{P}_t} \mathcal{J}(\mathbf{P}_t) = \big\langle -\mathbf{g}_\text{new},\, \nabla_{\boldsymbol{\theta}} \mathcal{L}_\text{pre}\big(\boldsymbol{\theta};\, \mathbf{S}(\mathbf{P}_t)\big) \big\rangle = -\mathbf{g}_\text{new}^\top \nabla_{\boldsymbol{\theta}} \mathcal{L}_\text{pre}. \nonumber
\end{equation}
To optimize this via gradient ascent, we must compute the derivative of $\mathcal{J}(\mathbf{P}_t)$ with respect to the prompt parameters $\mathbf{P}_t$. Since the new task gradient $\mathbf{g}_\text{new}$ is computed on $\mathcal{T}_t$ and is strictly independent of $\mathbf{P}_t$, it acts as a constant during this differentiation. Applying the chain rule yields:
\begin{equation}
    \nabla_{\mathbf{P}_t} \mathcal{J}(\mathbf{P}_t) = \nabla_{\mathbf{P}_t} \Big[ -\mathbf{g}_\text{new}^\top \nabla_{\boldsymbol{\theta}} \mathcal{L}_\text{pre} \Big] = - \Big( \nabla^2_{\mathbf{P}_t, \boldsymbol{\theta}} \mathcal{L}_\text{pre} \Big) \mathbf{g}_\text{new}. \nonumber
\end{equation}
The term $\nabla^2_{\mathbf{P}_t, \boldsymbol{\theta}} \mathcal{L}_\text{pre}$ is a \textit{mixed second-order derivative matrix}, \ie a Hessian cross-derivative. Computing a Hessian-vector product through the massive layers of an LLM requires unrolling the entire computational graph of the first gradient backward. This triggers an $\mathcal{O}(N^2)$ memory explosion and massive computational latency, making it absolutely intractable for LLMs.

\noindent \textbf{First-order resolution of \baby.}
Instead of differentiating the inner product, \baby employs a \textit{virtual step approximation}. According to the first-order Taylor expansion of $\mathcal{L}_\text{pre}$ around the current weights $\boldsymbol{\theta}_t$ with a step size $\alpha$:
\begin{equation}
    \mathcal{L}_\text{pre}(\boldsymbol{\theta}_t - \alpha \mathbf{g}_\text{new}) \approx \mathcal{L}_\text{pre}(\boldsymbol{\theta}_t) - \alpha \mathbf{g}_\text{new}^\top \nabla_{\boldsymbol{\theta}} \mathcal{L}_\text{pre}(\boldsymbol{\theta}_t). \nonumber
\end{equation}
Notice that the term $-\mathbf{g}_\text{new}^\top \nabla_{\boldsymbol{\theta}} \mathcal{L}_\text{pre}$ is exactly our desired gradient conflict objective $\mathcal{J}(\mathbf{P}_t)$. By defining the virtual weights as $\boldsymbol{\theta}_\text{virt} = \boldsymbol{\theta}_t - \alpha \mathbf{g}_\text{new}$, the maximization problem becomes:
\begin{equation}
    \mathop{\boldsymbol{\max}}\limits_{\mathbf{P}_t} \mathcal{L}_\text{pre}\big(\boldsymbol{\theta}_\text{virt};\, \mathbf{S}(\mathbf{P}_t)\big). \nonumber
\end{equation}
To optimize this new objective, we simply take the derivative \wrt $\mathbf{P}_t$:
\begin{equation}
    \nabla_{\mathbf{P}_t} \mathcal{L}_\text{pre}\big(\boldsymbol{\theta}_\text{virt};\, \mathbf{S}(\mathbf{P}_t)\big). \nonumber
\end{equation}
Because $\boldsymbol{\theta}_\text{virt}$ is treated as a fixed parameter matrix during this step, this operation is a \textit{standard, first-order backpropagation}. We completely bypass the Hessian matrix, reducing the spatial and temporal complexity from $\mathcal{O}(N^2)$ to $\mathcal{O}(N)$, rendering the adversarial training highly scalable.

\subsection{Time Complexity Compared with Standard LoRA Baselines}
\label{subsec:time_complexity}

We now quantify the exact computational overhead introduced by \baby compared to standard LoRA fine-tuning. We define the time complexity of a standard LLM forward pass as $\mathcal{O}(\text{FP}_\text{LLM})$, a backward pass to update LoRA weights as $\mathcal{O}(\text{BP}_\text{LoRA})$, and a backward pass to update the soft prompt as $\mathcal{O}(\text{BP}_\text{Prompt})$. Note that $\text{BP}_\text{LoRA} \approx \text{BP}_\text{Prompt} \ll \text{BP}_\text{Full\_Model}$.

In a standard LoRA fine-tuning iteration, the model processes a batch of new task data, computes the loss, and backpropagates to update the LoRA parameters, \ie {\color{Blue} $\mathcal{O}(\text{FP}_\text{LLM}) + \mathcal{O}(\text{BP}_\text{LoRA})$}.

Turn to analyze \baby, a single iteration of \baby performs the following sequential steps:

\hangingpar{
\textbf{1. New task gradient (\textit{Step 1}):} Compute $\mathbf{g}_\text{new}$ on $\mathcal{T}_t$, \ie $\mathcal{O}(\text{FP}_\text{LLM}) + \mathcal{O}(\text{BP}_\text{LoRA})$;
}
\hangingpar{
\textbf{2. Pre-training gradient estimation (\textit{Sec.~\ref{subsec:generative_replay}}):} Auto-regressively generating a soft sequence of length $M$ using Gumbel-Softmax requires generating $M$ tokens iteratively, so the cost is $M \times \mathcal{O}(\text{FP}_\text{LLM})$;
}
\hangingpar{
\textbf{3. Adversarial prompt update (\textit{Step 1}):} Evaluate $\mathcal{L}_\text{pre}$ using the generated sequence on $\boldsymbol{\theta}_\text{virt}$, and backpropagate to update $\mathbf{P}_t$, \ie $\mathcal{O}(\text{FP}_\text{LLM}) + \mathcal{O}(\text{BP}_\text{Prompt})$;
}
\hangingpar{
\textbf{4. Protection gradient calculation (\textit{Step 1}):} Evaluate $\mathcal{L}_\text{pre}$ again on the true weights $\boldsymbol{\theta}_t$ to obtain $\mathbf{g}_\text{pre}^*$, \ie $\mathcal{O}(\text{FP}_\text{LLM}) + \mathcal{O}(\text{BP}_\text{LoRA})$;
}
\hangingpar{
\textbf{5. Pareto orthogonal projection (\textit{Steps 2 \& 3}):} Project the gradients using the historical memory matrix $\mathbf{M}_\text{new}$, which contains $t-1$ vectors of dimension $D_\text{LoRA}$. This involves basic matrix-vector multiplications on the parameter space, which is negligible compared to passing tensors through the LLM layers. Therefore, its cost is $\mathcal{O}(D_\text{LoRA} \cdot t) \approx \mathcal{O}(1)$ relative to LLM passes.
}
Summing the costs, a single step of \baby requires {\color{Blue} $\mathcal{O}\big( (M+3)\text{FP}_\text{LLM} + 2\text{BP}_\text{LoRA} + \text{BP}_\text{Prompt} \big)$}.
Because the length of the surrogate sequence $M$ is typically a small constant, the overall complexity of \baby scales \textit{linearly} with the LLM size, identically to standard LoRA. We successfully mitigate the catastrophic forgetting by introducing only a \textit{small constant-factor computational overhead}, completely avoiding the exponential $\mathcal{O}(N^2)$ scaling that traditionally plagues orthogonal gradient alignment and adversarial generation in LLMs.

\subsection{Empirical Computational Budget}
\label{subsec:empirical_budget}

In Table~\ref{budget}, we present an empirical comparison of the computational time required by our proposed method, \baby, against standard LoRA and other advanced continual LLM fine-tuning baselines, including LoRAMoE \citep{dou2023loramoe} and CLoRA \citep{lu2025controlled}. All experiments are measured in running minutes using an 8$\times$GPUs distributed training setup across various representative LLM architectures (Qwen3, Llama3, and Gemma2 families). 
As expected, standard LoRA serves as the lower bound for training time (averaging 60 minutes, denoted as $1\times$) due to its straightforward low-rank weight injection. However, methods designed to enhance model capacity or mitigate catastrophic forgetting inherently introduce substantial computational overhead. For instance, CLoRA aims to reduce output variations and preserve previous knowledge by applying subspace regularization, which imposes strict mathematical constraints on the direction of the updating matrix's null space. While effective for continual learning, this regularization mechanism significantly burdens the training process, resulting in an average computational cost of $2.6\times$ compared to standard LoRA. Similarly, LoRAMoE incurs the highest computational budget (averaging $3.2\times$) due to the complex routing mechanisms and multiple expert networks involved during the forward and backward passes.

In contrast, our proposed \baby demonstrates superior computational efficiency among the advanced fine-tuning methods. Across all evaluated model scales, \baby consistently requires considerably less training time than both CLoRA and LoRAMoE. On average, \baby only takes $2.0\times$ the time of standard LoRA, saving roughly 23\% and 37\% of the training time compared to CLoRA and LoRAMoE, respectively. Notably, the efficiency of \baby becomes even more pronounced on relatively compact models; for example, on Llama3-3B and Gemma2-2B, the relative computational cost drops to just $1.8\times$ and $1.5\times$. 
These results indicate that \baby successfully bypasses the severe computational bottlenecks typically associated with strict orthogonal gradient projection (as seen in CLoRA) or dense mixture-of-experts routing (as in LoRAMoE). Consequently, \baby offers a highly practical and scalable solution, striking an optimal balance between advanced model adaptability and empirical training efficiency.

\begin{table}[t]
\centering
\renewcommand\arraystretch{1.1}
\caption{Empirical computational time (minutes) on 8$\times$GPUs.}
\label{budget}
\small
\setlength{\tabcolsep}{5pt}{
    \begin{tabular}{m{1.75cm}<{\centering}m{1.4cm}<{\centering}m{1.55cm}<{\centering}m{1.55cm}<{\centering}m{1.5cm}<{\centering}m{1.5cm}<{\centering}m{1.5cm}<{\centering}}
    \toprule
    LLM & LoRA & LoRAMoE \textit{(\#expert=4)} & LoRAMoE \textit{(\#expert=8)} & OLoRA & CLoRA & \baby \\
    \hline
    Qwen3-8B & 88 (1$\times$) & 223 (2.5$\times$) & 225 (2.5$\times$) & 237 (2.7$\times$) & 191 (2.2$\times$) & 162 (1.8$\times$) \\
    Qwen3-4B & 62 (1$\times$) & 194 (3.1$\times$) & 195 (3.1$\times$) & 204 (3.3$\times$) & 155 (2.5$\times$) & 129 (2.1$\times$) \\
    Llama3-8B & 60 (1$\times$) & 197 (3.3$\times$) & 199 (3.3$\times$) & 211 (3.5$\times$) & 156 (2.6$\times$) & 130 (2.1$\times$) \\
    Llama3-3B & 35 (1$\times$) & 144 (4.1$\times$) & 144 (4.1$\times$) & 151 (4.3$\times$) & 114 (3.2$\times$) & 65 (1.8$\times$) \\
    Gemma2-9B & 76 (1$\times$) & 260 (3.4$\times$) & 262 (3.4$\times$) & 271 (3.5$\times$) & 210 (2.7$\times$) & 178 (2.3$\times$) \\
    Gemma2-2B & 44 (1$\times$) & 140 (3.2$\times$) & 141 (3.2$\times$) & 149 (3.4$\times$) & 116 (2.6$\times$) & 69 (1.5$\times$) \\
    \hline
    Average & 60 (1$\times$) & 193 (3.2$\times$) & 194 (3.2$\times$) & 203 (3.4$\times$) & 157 (2.6$\times$) & 122 (2.0$\times$) \\
    \bottomrule
    \end{tabular} }
\end{table}

\section{Algorithm and Notations} \label{sec.appendix.alg}

In this section, we summarize the notations used in our proposed method and provide the complete algorithmic workflow of \baby. 
Table~\ref{tab:notations} summarizes the primary mathematical notations and their corresponding descriptions used throughout the formulation of our method. As analyzed in Sec.~\ref{subsec:efficient_solver}, directly solving the multi-objective optimization problem for continual fine-tuning is computationally intractable due to the natural adversariality and the requirement of second-order Hessian matrices. To address this, \baby elegantly decouples the complex optimization into an efficient first-order bilevel training paradigm. The overall training procedure is outlined in Alg.~\ref{alg:baby}. For each new task, we first utilize a virtual step approximation to adversarially optimize the soft prompt, extracting the most vulnerable pre-training data distribution. Subsequently, we leverage null-space projection and Pareto-optimal orthogonal projection to eliminate gradient conflicts. This ensures that the newly updated large model efficiently acquires new task plasticity while strictly maintaining general-purpose knowledge stability and retaining previous downstream task abilities.

\begin{table}[t]
    \centering
    \renewcommand\arraystretch{1.15}
    \caption{Summary of primary notations used in \baby.}
    \label{tab:notations}
    \small
    \setlength{\tabcolsep}{8pt}{
    \begin{tabular}{cl}
        \toprule
        \textbf{Notation} & \textbf{Description} \\
        \midrule
        $\mathcal{F}_{\boldsymbol{\theta}}, \boldsymbol{\theta}$ & Frozen pre-training LLM and trainable LLM weights \\
        $\mathcal{T}_t, \mathcal{T}_0$ & $t$-th new downstream task and original pre-training distribution \\
        $\mathbf{P}_t, \mathbf{S}(\mathbf{P}_t)$ & Continuous soft prompt and generated surrogate sequence \\
        $\mathcal{L}_\text{new}, \mathcal{L}_\text{pre}$ & Plasticity and stability objective functions \\
        $\mathbf{g}_\text{new}, \mathbf{g}_\text{pre}$ & Task-specific and protection gradients \wrt $\boldsymbol{\theta}$ \\
        $\boldsymbol{\theta}_\text{virt}$ & Virtually updated weights simulating new task impact \\
        $\mathcal{M}_\text{new}, \boldsymbol{\Pi}_\text{new}$ & Historical gradient memory bank and its null-space projection matrix \\
        $\tilde{\mathbf{g}}_\text{new}, \tilde{\mathbf{g}}_\text{pre}$ & Gradients projected into historically safe null space \\
        $\tilde{\mathbf{g}}_\text{new}^*$ & Pareto-optimal modified new task gradient \\
        $\eta, \alpha, \lambda$ & learning rate, virtual step size, and orthogonal penalty coefficient \\
        \bottomrule
    \end{tabular}}
\end{table}

\begin{algorithm}[t]
\caption{Training Process of \baby}
\label{alg:baby}
\begin{algorithmic}[1]
\REQUIRE Frozen pre-training LLM $\mathcal{F}_{\boldsymbol{\theta}}$, task sequence $\{\mathcal{T}_1, \dots, \mathcal{T}_T\}$, learning rates $\eta, \alpha$, penalty coefficient $\lambda$.
\ENSURE Optimized LLM parameters $\boldsymbol{\theta}$.

\STATE Initialize trainable LLM parameters $\boldsymbol{\theta}$ with $\boldsymbol{\theta}^0$.
\STATE Initialize historical gradient memory sets $\mathcal{M}_\text{new} \leftarrow \varnothing$, $\mathcal{M}_\text{pre} \leftarrow \varnothing$.

\FOR{each new task $t = 1, 2, \dots, T$}
    \STATE Initialize the learnable soft prompt $\mathbf{P}_t$.
    \FOR{each training iteration}
        \STATE Sample a mini-batch of data from $\mathcal{T}_t$.
        \STATE Compute new task gradient $\mathbf{g}_\text{new} = \nabla_{\boldsymbol{\theta}} \mathcal{L}_\text{new}$.
        
        \STATE \textit{\color{Blue} $\triangleright$ Step 1: Vulnerability \& pre-training gradient estimation}
        \STATE Perform virtual update $\boldsymbol{\theta}_\text{virt}$ via Eq.~(\ref{eq:virtual_step}).
        \STATE Optimize $\mathbf{P}_t$ by maximizing Eq.~(\ref{eq:constraint_adv}) to generate $\mathbf{S}(\mathbf{P}_t^*)$.
        \STATE Compute protection gradient $\mathbf{g}_\text{pre}^* = \nabla_{\boldsymbol{\theta}} \mathcal{L}_\text{pre}(\boldsymbol{\theta}; \mathbf{S}(\mathbf{P}_t^*))$.
        
        \STATE \textit{\color{Blue} $\triangleright$ Step 2: Null-space projection}
        \STATE Construct projection matrix $\boldsymbol{\Pi}_\text{new}$ via Eq.~(\ref{eq:null_space}) using $\mathcal{M}_\text{new}$. 
        \STATE Obtain safe gradients $\tilde{\mathbf{g}}_\text{new} = \boldsymbol{\Pi}_\text{new} \mathbf{g}_\text{new}$ and $\tilde{\mathbf{g}}_\text{pre} = \boldsymbol{\Pi}_\text{new} \mathbf{g}_\text{pre}^*$.
        
        \STATE \textit{\color{Blue} $\triangleright$ Step 3: Pareto orthogonal optimization}
        \IF{$\langle \tilde{\mathbf{g}}_\text{new}, \tilde{\mathbf{g}}_\text{pre} \rangle < 0$}
            \STATE Compute interference-free gradient $\tilde{\mathbf{g}}_\text{new}^*$ via Pareto projection in Eq.~(\ref{eq:pareto_projection}).
        \ELSE
            \STATE Set $\tilde{\mathbf{g}}_\text{new}^* = \tilde{\mathbf{g}}_\text{new}$.
        \ENDIF
        
        \STATE \textit{\color{Blue} $\triangleright$ Step 4: Parameter update}
        \STATE Update LLM parameters $\boldsymbol{\theta}$ following the direction in Eq.~(\ref{eq:final_update}). 
    \ENDFOR
    \STATE Update memory: $\mathcal{M}_\text{new} \leftarrow \mathcal{M}_\text{new} \cup \{\mathbf{g}_\text{new}\}$, $\mathcal{M}_\text{pre} \leftarrow \mathcal{M}_\text{pre} \cup \{\mathbf{g}_\text{pre}^*\}$. 
\ENDFOR
\RETURN $\boldsymbol{\theta}$
\end{algorithmic}
\end{algorithm}

\section{Experimental Setups} \label{sec:appendix.setup}

To rigorously evaluate \baby under a continual fine-tuning setting, we conduct experiments on a broad suite of multi-task benchmarks and multiple open-source pre-training LLMs. In this section, we describe the pre-training LLMs, datasets, baseline methods, evaluation metrics, and implementation details used throughout our experiments.

\subsection{Pre-training LLMs}\label{sec.appendix.detail.model}
\vspace{-2.5pt}

To comprehensively evaluate the effectiveness of our method, we conduct experiments on three prominent families of open-weight decoder-only LLMs, selecting two parameter scales for each family: \textbf{Qwen3} (\textbf{4B} and \textbf{8B}) \citep{qwen2025qwen3}, \textbf{Llama3} (\textbf{3B} and \textbf{8B}) \citep{llama3modelcard}, and \textbf{Gemma2} (\textbf{2B} and \textbf{9B}) \citep{gemma2024gemma}. These LLM families provide a diverse and representative testbed spanning different pre-training corpora, architectural variations, and capacity regimes:
\begin{itemize}
    \item \textit{The Qwen3 family} \citep{qwen2025qwen3} is renowned for its exceptional multilingual capabilities and robust alignment with human instructions. We include the 4B and 8B variants to investigate continual learning dynamics, which distill from their larger-scale variant, \eg Qwen3-32B, by the off-policy and on-policy distillation.
    \item \textit{The Llama3 family} \citep{llama3modelcard} is pre-trained on a massive and highly curated corpus of over 9 trillion tokens. It features an optimized Transformer architecture with grouped-query attention and an expanded vocabulary size. We utilize the 3B and 8B variants to represent highly capable models at accessible scales.
    \item \textit{The Gemma2 family} \citep{gemma2024gemma} introduces architectural innovations such as interleaved local and global attention mechanisms and logit soft-capping. The 2B and 9B models are selected to evaluate our framework's stability on highly optimized, dense architectures.
\end{itemize}
By evaluating across these distinct families and scales, we aim to demonstrate that our continual fine-tuning framework is robust and generally applicable regardless of the underlying LLM architecture.

\begin{table}[t]
\centering
\renewcommand\arraystretch{1.15}
\setlength{\tabcolsep}{4pt}
\caption{Overview of selected multi-task datasets from \textit{SuperNI} \citep{wang2022super}.}
\label{dataset_new}
\small
    \begin{tabular}{m{1.0cm}<{\centering}m{5.8cm}<{\centering}m{2.9cm}<{\centering}m{0.9cm}<{\centering}m{0.9cm}<{\centering}}
    \toprule
    ID & Dataset & Task Type & Train & Eval \\
    \midrule
    % Text Summarization
    \rowcolor{gray!10} t1290 & xsum\_summarization & text summarization & 1,000 & 100 \\
    \rowcolor{gray!10} t511  & reddit\_tifu\_long\_text\_summarization & text summarization & 1,000 & 100 \\
    \rowcolor{gray!10} t1572 & samsum\_summary & text summarization & 160 & 20 \\
    \hline
    % Dialogue Generation
    t1729 & personachat\_generate\_next & dialogue generation & 1,000 & 100 \\
    t639  & multi\_woz\_user\_utterance\_generation & dialogue generation & 142 & 18 \\
    t1590 & diplomacy\_text\_generation & dialogue generation & 126 & 16 \\
    \hline
    % Sentiment Analysis
    \rowcolor{gray!10} t1687 & sentiment140\_classification & sentiment analysis & 1,000 & 100 \\
    \rowcolor{gray!10} t363  & sst2\_polarity\_classification & sentiment analysis & 1,000 & 100 \\
    \rowcolor{gray!10} t875  & emotion\_classification & sentiment analysis & 1,000 & 100 \\
    \hline
    % Information Extraction
    t748  & glucose\_reverse\_cause\_event\_detection & information extraction & 1,000 & 100 \\
    t1510 & evalution\_relation\_extraction & information extraction & 1,000 & 100 \\
    t181  & outcome\_extraction & information extraction & 338 & 43 \\
    \hline
    % Question Answering
    \rowcolor{gray!10} t591 & sciq\_answer\_generation & question answering & 1,000 & 100 \\
    \rowcolor{gray!10} t002 & quoref\_answer\_generation & question answering & 1,000 & 100 \\
    \rowcolor{gray!10} t073 & commonsenseqa\_answer\_generation & question answering & 975 & 100 \\
    \bottomrule
    \end{tabular}
\end{table}

\begin{table}[t]
\centering
\renewcommand\arraystretch{1.15}
\caption{Three distinct task orderings under the continual LLM fine-tuning.}
\label{taskorder}
\small
\setlength{\tabcolsep}{5pt}{
    \begin{tabular}{m{0.5cm}<{\centering}m{12cm}<{\centering}}
    \toprule
    Seed & Task Ordering \\
    \hline
    \rowcolor{gray!10} 1 & $\text{t591} \rightarrow \text{t073} \rightarrow \text{t1687} \rightarrow \text{t875} \rightarrow \text{t1572} \rightarrow \text{t639} \rightarrow \text{t1510} \rightarrow \text{t1590} \rightarrow \text{t511} \rightarrow \text{t181} \rightarrow \text{t363} \rightarrow \text{t1729} \rightarrow \text{t002} \rightarrow \text{t1290} \rightarrow \text{t748}$ \\
    \hline
    2 & $\text{t1290} \rightarrow \text{t1510} \rightarrow \text{t1572} \rightarrow \text{t002} \rightarrow \text{t073} \rightarrow \text{t511} \rightarrow \text{t1590} \rightarrow \text{t363} \rightarrow \text{t748} \rightarrow \text{t639} \rightarrow \text{t181} \rightarrow \text{t1729} \rightarrow \text{t1687} \rightarrow \text{t591} \rightarrow \text{t875}$ \\
    \hline
    \rowcolor{gray!10} 3 & $\text{t002} \rightarrow \text{t073} \rightarrow \text{t1572} \rightarrow \text{t181} \rightarrow \text{t591} \rightarrow \text{t1687} \rightarrow \text{t363} \rightarrow \text{t1729} \rightarrow \text{t511} \rightarrow \text{t875} \rightarrow \text{t639} \rightarrow \text{t748} \rightarrow \text{t1590} \rightarrow \text{t1290} \rightarrow \text{t1510}$ \\
    \bottomrule
    \end{tabular} }
\end{table}

\subsection{Benchmark Datasets} \label{sec.appendix.detail.dataset}
\label{sec:datasets}

We evaluate our continual LLM fine-tuning framework on two complementary benchmarks, \textit{SuperNI} and \textit{MMLU}, aiming to jointly assess task-specific adaptation and general-purpose knowledge preservation under a continual training stream. For continual task learning, we adopt a subset of the \textit{SuperNI} benchmark \citep{wang2022super}, consisting of 15 diverse tasks spanning five categories, including question answering, information extraction, sentiment analysis, text summarization, and dialogue generation, as summarized in Table~\ref{dataset_new}. Each task is associated with its own training and evaluation splits, and the LLM is trained sequentially over the task stream, where each task is observed only once. After completing the training on each task, we evaluate the LLM on the evaluation sets of all tasks, enabling a unified assessment of their accuracy and forgetting rate on previously learned tasks. To mitigate the bias induced by a specific task sequence, we further consider three distinct task orderings generated from different random seeds, as shown in Table~\ref{taskorder}, and conduct all experiments independently on each ordering to ensure robustness against ordering effects. 

\begin{wraptable}{r}{0.42\textwidth}
\vspace{-5pt}
\centering
\renewcommand{\arraystretch}{1.15}
\setlength{\tabcolsep}{4pt}
\caption{Statistic of \textit{MMLU} subjects.}
\label{dataset_mmlu}
\small
    \begin{tabular}{m{2.6cm}<{\centering}m{1.5cm}<{\centering}m{0.7cm}<{\centering}}
    \toprule
    Subject & Category & \#Eval \\
    \midrule
    \rowcolor{gray!10} anatomy & other & 135 \\
    \rowcolor{gray!10} global\_facts & other & 100 \\
    \rowcolor{gray!10} marketing & other & 234 \\
    philosophy & humanities & 311 \\
    formal\_logic & humanities & 283 \\
    prehistory & humanities & 324 \\
    \rowcolor{gray!10} high\_school\_math & STEM & 270 \\
    \rowcolor{gray!10} high\_school\_physics & STEM & 154 \\
    \rowcolor{gray!10} high\_school\_CS & STEM & 287 \\
    \bottomrule
    \end{tabular}
\vspace{-5pt}
\end{wraptable}
In addition to task-specific performance, we evaluate the LLM's general-purpose knowledge using a subset of the \textit{MMLU} benchmark \citep{hendrycks2021measuring}, which covers a broad range of domains including STEM, humanities, and other disciplines, as listed in Table~\ref{dataset_mmlu}. Following the standard \textit{MMLU} protocol, the development split is used to construct 5-shot contexts, while performance is measured on the test split. Importantly, consistent with our continual learning setup, we interleave evaluation with training by testing the LLM on the full \textit{MMLU} benchmark after each task update. This design allows us to track the evolution of general-purpose knowledge throughout the continual fine-tuning process, thereby providing a comprehensive view of the trade-off between task specialization and generalization in LLMs.

\subsection{Baselines Methods}
We compare our method with several representative continual LLM fine-tuning approaches as follows:
\begin{itemize}
    \item \textbf{LoRA} \citep{hu2022lora} is a widely adopted method that isolates and reparameterizes the weight updates of a pre-trained LLM into low-rank matrices. By freezing the backbone parameters and only optimizing the low-rank adapters, LoRA significantly reduces the number of trainable parameters.
    \item \textbf{LoRAMoE} \citep{dou2023loramoe} extends LoRA by incorporating a mixture-of-experts mechanism into the low-rank adaptation process. Specifically, multiple LoRA experts are maintained, and a routing strategy dynamically selects or combines these experts for each input, thereby improving model capacity and adaptability across tasks.
    \item \textbf{O-LoRA} \citep{wang2023orthogonal} introduces orthogonality constraints on the LoRA update matrices to mitigate interference between tasks. By encouraging the updates of different tasks to lie in orthogonal subspaces, O-LoRA reduces catastrophic forgetting while preserving the efficiency of low-rank adaptation.
    \item \textbf{GainLoRA} \citep{liang2025gated} further enhances the LoRA-based continual learning paradigm by introducing a gated integration mechanism over multiple LoRA branches. For each new task, a new LoRA branch is added, and learnable gating modules are used to control the contribution of each branch. This design effectively suppresses the influence of newly learned parameters on previous tasks, thereby alleviating forgetting without requiring task identity during inference.
    \item \textbf{CLoRA} \cite{lu2025controlled} is a learning-based method that imposes subspace regularization on LoRA updates. Instead of only constraining the rank of the update, CLoRA further restricts the direction of the update matrix by enforcing orthogonality with respect to a predefined subspace. This strategy reduces the magnitude of output perturbations while maintaining sufficient model capacity, achieving a better trade-off between adaptation and forgetting.
\end{itemize}
These baselines collectively cover standard LoRA fine-tuning variants and state-of-the-art continual fine-tuning strategies, providing a comprehensive comparison for evaluating our method.

\subsection{Evaluation Metrics}\label{sec.appendix.detail.metrics}

To rigorously assess both task-specific adaptation and the preservation of previously acquired general-purpose knowledge during continual fine-tuning, we define specific performance and forgetting metrics for the \textit{SuperNI} and \textit{MMLU} benchmarks.

\noindent
\textbf{\textit{SuperNI} metrics.} 
Given that \textit{SuperNI} comprises diverse generation tasks, we utilize the standard \textsc{Rouge} score (specifically \textsc{Rouge-L}) to measure the generation quality by computing the longest common subsequence overlap between the LLM's prediction and the ground-truth reference.
To evaluate the stability of the LLM against catastrophic forgetting on sequential tasks, we introduce the \textit{forgetting rate} (Fgt.).
For a continually fine-tuned LLM, let $R_{t,t}$ denote the \textsc{Rouge} score evaluated on task $t$'s test set immediately after the LLM finishes training on task $t$. Let $R_{T,t}$ denote the \textsc{Rouge} score on task $t$ after the LLM has sequentially finished training on all $T$ tasks. The forgetting rate for task $t$ is defined as the performance drop between these two stages:
\begin{equation}
    \mathrm{FR}_{t} = R_{t,t} - R_{T,t}. \nonumber
\end{equation}
The overall forgetting rate on \textit{SuperNI} is calculated as the average of $\mathrm{FR}_{t}$ across all $T$ learned tasks: $\mathrm{FR}_{\textit{SuperNI}} = \frac{1}{T} \sum_{t=1}^{T} \mathrm{FR}_{t}$. A lower forgetting rate indicates that the model better retains the knowledge of previously learned tasks.

\noindent
\textbf{\textit{MMLU} metrics.} 
Since \textit{MMLU} is a multi-choice selection dataset designed to test general-purpose knowledge, we use \textit{accuracy} as the primary metric. The overall accuracy is computed by pooling all evaluated questions across the selected subjects. 
To measure the degradation of the LLM's pre-trained knowledge after adapting to the sequence of downstream tasks, we define the \textit{MMLU} forgetting rate. Let $A_{0}$ denote the \textit{MMLU} accuracy of the original pre-training LLM before any continual fine-tuning. Let $A_{T}$ denote the MMLU accuracy of the LLM after it has been trained on all $T$ tasks in the continual learning stream. The general-purpose knowledge forgetting rate is defined as:
\begin{equation}
    \mathrm{FR}_{\textit{MMLU}} = A_{0} - A_{T}. \nonumber
\end{equation}
This metric explicitly quantifies the trade-off between acquiring new task-specific skills and preserving the pre-training model's broad, general-purpose knowledge.

\subsection{Implementation Details} \label{app:implementation}

Our experiments are implemented in PyTorch based on Hugging Face \texttt{Transformers} and \texttt{PEFT}. We fine-tune the backbone LLMs with LoRA while keeping the original pre-training LLM parameters frozen. During fine-tuning, gradient checkpointing is enabled to lower activation memory consumption, and the input sequence length is truncated to 1024 tokens with the maximum generation length set to 50 tokens. We optimize the trainable parameters with AdamW using an initial learning rate of $2 \times 10^{-4}$. To ensure reproducibility, the task order is shuffled once per run under a fixed random seed, and evaluation is performed after each task to track both task acquisition and forgetting over the course of continual learning.
For \textit{MMLU} evaluation, we follow the standard 5-shot protocol and set the maximum context length to 2048 tokens. If the composed prompt exceeds the context budget, the number of demonstrations is reduced until the input fits within the limit. During inference, the LLM predicts among the four candidate answers by computing the logits at the final token position and selecting the option with the highest probability. We use batch size 1 for MMLU to preserve the integrity of the few-shot prompt, and this evaluation is run in the same continual-learning pipeline as \textit{SuperNI} to obtain a consistent estimate of both in-domain adaptation and out-of-domain retention.

\noindent
\textbf{Stable optimization of pre-training gradient estimation.}
As introduced in Sec.~\ref{subsec:generative_replay}, \baby employs the Gumbel-Softmax relaxation to maintain a differentiable computational graph across auto-regressive decoding steps. However, applying Gumbel-Softmax over the entire vocabulary $\mathcal{V}$ (typically over 30k tokens for modern LLMs) across a long sequence can lead to severe memory consumption and gradient instability (\eg vanishing or exploding gradients).
To address this, we introduce two engineering adaptations. First, we restrict the surrogate sequence length to a moderate size (\eg $M=16$ or $32$). Empirical observations suggest that the model does not need to generate excessively long context to expose general-purpose knowledge vulnerabilities; a short, highly-targeted surrogate sequence $\mathbf{S}(\mathbf{P}_t)$ optimized via the virtual step is entirely sufficient to anchor the general-purpose knowledge boundaries. Second, to alleviate the memory overhead during the softmax expectation, we apply a dynamic vocabulary truncation strategy: at each generation step $m$, we only compute the Gumbel-Softmax over the top-50 highest probability tokens in $\mathbf{z}_m$, effectively zeroing out the long-tail noise and drastically reducing the size of the differentiable computational graph.

\section{Performance Dynamics on \textit{SuperNI}}

In Tables~\ref{oursseed1qwen3} and \ref{gainloraseed1qwen3}, we compare the performance dynamics of our approach against the baseline methods across each task in \textit{SuperNI}, recorded sequentially after the training of every task. Our results demonstrate that our method outperforms the baselines in the vast majority of scenarios.

\definecolor{tomato}{rgb}{0.9, 0.3, 0.3}
% ================== Qwen3-4B ours SuperNI ====================
\begin{table*}[t]
\centering
\caption{Performance dynamics during continual fine-tuning, using the \textbf{Qwen3-4B} model with the \textbf{\baby} method (\textsc{Rouge} = 50.7, Forget Rate = 6.0).}
\renewcommand\arraystretch{0.97}
\setlength{\tabcolsep}{4pt}
\footnotesize
\label{oursseed1qwen3}
\begin{tabular}{|l|*{17}{c|}}
\hline
\multicolumn{2}{|c|}{} & \multicolumn{15}{c|}{Evaluation Task ID} \\
\cline{3-17}
\multicolumn{2}{|c|}{} & 591 & 073 & 1687 & 875 & 1572 & 639 & 1510 & 1590 & 511 & 181 & 363 & 1729 & 002 & 1290 & 748 \\
\hline
\multirow{15}{*}{\rotatebox{90}{Training Task ID}} 
& 591 & \cellcolor{tomato!67.2}67.2 & \cellcolor{tomato!80.1}80.1 & \cellcolor{tomato!42.5}42.5 & \cellcolor{tomato!41.0}41.0 & \cellcolor{tomato!32.6}32.6 & \cellcolor{tomato!8.2}8.2 & \cellcolor{tomato!24.2}24.2 & \cellcolor{tomato!11.2}11.2 & \cellcolor{tomato!14.8}14.8 & \cellcolor{tomato!13.9}13.9 & \cellcolor{tomato!60.6}60.6 & \cellcolor{tomato!10.6}10.6 & \cellcolor{tomato!77.1}77.1 & \cellcolor{tomato!18.8}18.8 & \cellcolor{tomato!42.9}42.9 \\
& 073 & \cellcolor{tomato!68.9}68.9 & \cellcolor{tomato!81.0}81.0 & \cellcolor{tomato!61.4}61.4 & \cellcolor{tomato!45.0}45.0 & \cellcolor{tomato!32.0}32.0 & \cellcolor{tomato!7.2}7.2 & \cellcolor{tomato!15.3}15.3 & \cellcolor{tomato!13.7}13.7 & \cellcolor{tomato!14.3}14.3 & \cellcolor{tomato!13.8}13.8 & \cellcolor{tomato!92.0}92.0 & \cellcolor{tomato!10.5}10.5 & \cellcolor{tomato!78.2}78.2 & \cellcolor{tomato!19.1}19.1 & \cellcolor{tomato!42.8}42.8 \\
& 1687 & \cellcolor{tomato!66.8}66.8 & \cellcolor{tomato!77.0}77.0 & \cellcolor{tomato!86.0}86.0 & \cellcolor{tomato!42.0}42.0 & \cellcolor{tomato!32.4}32.4 & \cellcolor{tomato!7.3}7.3 & \cellcolor{tomato!54.4}54.4 & \cellcolor{tomato!13.6}13.6 & \cellcolor{tomato!14.4}14.4 & \cellcolor{tomato!11.6}11.6 & \cellcolor{tomato!91.0}91.0 & \cellcolor{tomato!10.9}10.9 & \cellcolor{tomato!77.6}77.6 & \cellcolor{tomato!18.7}18.7 & \cellcolor{tomato!43.9}43.9 \\
& 875 & \cellcolor{tomato!62.9}62.9 & \cellcolor{tomato!77.0}77.0 & \cellcolor{tomato!85.0}85.0 & \cellcolor{tomato!81.0}81.0 & \cellcolor{tomato!35.2}35.2 & \cellcolor{tomato!6.6}6.6 & \cellcolor{tomato!74.8}74.8 & \cellcolor{tomato!12.0}12.0 & \cellcolor{tomato!13.4}13.4 & \cellcolor{tomato!23.8}23.8 & \cellcolor{tomato!84.0}84.0 & \cellcolor{tomato!10.2}10.2 & \cellcolor{tomato!73.8}73.8 & \cellcolor{tomato!20.7}20.7 & \cellcolor{tomato!43.3}43.3 \\
& 1572 & \cellcolor{tomato!61.0}61.0 & \cellcolor{tomato!67.0}67.0 & \cellcolor{tomato!86.0}86.0 & \cellcolor{tomato!74.0}74.0 & \cellcolor{tomato!40.7}40.7 & \cellcolor{tomato!7.0}7.0 & \cellcolor{tomato!85.9}85.9 & \cellcolor{tomato!10.5}10.5 & \cellcolor{tomato!14.8}14.8 & \cellcolor{tomato!29.7}29.7 & \cellcolor{tomato!87.0}87.0 & \cellcolor{tomato!14.1}14.1 & \cellcolor{tomato!73.5}73.5 & \cellcolor{tomato!18.2}18.2 & \cellcolor{tomato!44.5}44.5 \\
& 639 & \cellcolor{tomato!52.4}52.4 & \cellcolor{tomato!59.0}59.0 & \cellcolor{tomato!82.0}82.0 & \cellcolor{tomato!78.0}78.0 & \cellcolor{tomato!35.1}35.1 & \cellcolor{tomato!11.6}11.6 & \cellcolor{tomato!73.5}73.5 & \cellcolor{tomato!12.3}12.3 & \cellcolor{tomato!14.1}14.1 & \cellcolor{tomato!29.3}29.3 & \cellcolor{tomato!84.0}84.0 & \cellcolor{tomato!12.8}12.8 & \cellcolor{tomato!69.4}69.4 & \cellcolor{tomato!18.9}18.9 & \cellcolor{tomato!41.1}41.1 \\
& 1510 & \cellcolor{tomato!60.7}60.7 & \cellcolor{tomato!65.0}65.0 & \cellcolor{tomato!82.0}82.0 & \cellcolor{tomato!78.0}78.0 & \cellcolor{tomato!38.1}38.1 & \cellcolor{tomato!8.5}8.5 & \cellcolor{tomato!100.0}100.0 & \cellcolor{tomato!14.0}14.0 & \cellcolor{tomato!14.5}14.5 & \cellcolor{tomato!36.3}36.3 & \cellcolor{tomato!86.0}86.0 & \cellcolor{tomato!13.7}13.7 & \cellcolor{tomato!73.1}73.1 & \cellcolor{tomato!19.6}19.6 & \cellcolor{tomato!43.0}43.0 \\
& 1590 & \cellcolor{tomato!54.1}54.1 & \cellcolor{tomato!70.0}70.0 & \cellcolor{tomato!82.0}82.0 & \cellcolor{tomato!74.0}74.0 & \cellcolor{tomato!28.8}28.8 & \cellcolor{tomato!4.1}4.1 & \cellcolor{tomato!99.7}99.7 & \cellcolor{tomato!10.8}10.8 & \cellcolor{tomato!13.7}13.7 & \cellcolor{tomato!29.3}29.3 & \cellcolor{tomato!80.0}80.0 & \cellcolor{tomato!10.0}10.0 & \cellcolor{tomato!69.6}69.6 & \cellcolor{tomato!18.3}18.3 & \cellcolor{tomato!41.8}41.8 \\
& 511 & \cellcolor{tomato!52.2}52.2 & \cellcolor{tomato!75.0}75.0 & \cellcolor{tomato!76.0}76.0 & \cellcolor{tomato!73.0}73.0 & \cellcolor{tomato!24.5}24.5 & \cellcolor{tomato!4.5}4.5 & \cellcolor{tomato!99.3}99.3 & \cellcolor{tomato!6.9}6.9 & \cellcolor{tomato!14.9}14.9 & \cellcolor{tomato!29.9}29.9 & \cellcolor{tomato!87.0}87.0 & \cellcolor{tomato!6.9}6.9 & \cellcolor{tomato!70.0}70.0 & \cellcolor{tomato!20.5}20.5 & \cellcolor{tomato!40.2}40.2 \\
& 181 & \cellcolor{tomato!60.0}60.0 & \cellcolor{tomato!70.0}70.0 & \cellcolor{tomato!83.0}83.0 & \cellcolor{tomato!70.0}70.0 & \cellcolor{tomato!18.9}18.9 & \cellcolor{tomato!1.6}1.6 & \cellcolor{tomato!99.7}99.7 & \cellcolor{tomato!7.8}7.8 & \cellcolor{tomato!15.2}15.2 & \cellcolor{tomato!71.3}71.3 & \cellcolor{tomato!84.0}84.0 & \cellcolor{tomato!6.7}6.7 & \cellcolor{tomato!72.6}72.6 & \cellcolor{tomato!18.9}18.9 & \cellcolor{tomato!41.4}41.4 \\
& 363 & \cellcolor{tomato!58.8}58.8 & \cellcolor{tomato!73.0}73.0 & \cellcolor{tomato!82.0}82.0 & \cellcolor{tomato!72.0}72.0 & \cellcolor{tomato!23.3}23.3 & \cellcolor{tomato!6.2}6.2 & \cellcolor{tomato!98.2}98.2 & \cellcolor{tomato!7.7}7.7 & \cellcolor{tomato!15.3}15.3 & \cellcolor{tomato!72.1}72.1 & \cellcolor{tomato!87.0}87.0 & \cellcolor{tomato!8.6}8.6 & \cellcolor{tomato!74.5}74.5 & \cellcolor{tomato!19.4}19.4 & \cellcolor{tomato!41.8}41.8 \\
& 1729 & \cellcolor{tomato!57.8}57.8 & \cellcolor{tomato!74.0}74.0 & \cellcolor{tomato!82.0}82.0 & \cellcolor{tomato!70.0}70.0 & \cellcolor{tomato!24.1}24.1 & \cellcolor{tomato!8.9}8.9 & \cellcolor{tomato!97.8}97.8 & \cellcolor{tomato!11.1}11.1 & \cellcolor{tomato!13.6}13.6 & \cellcolor{tomato!68.9}68.9 & \cellcolor{tomato!89.0}89.0 & \cellcolor{tomato!14.6}14.6 & \cellcolor{tomato!75.4}75.4 & \cellcolor{tomato!21.4}21.4 & \cellcolor{tomato!41.2}41.2 \\
& 002 & \cellcolor{tomato!59.8}59.8 & \cellcolor{tomato!71.0}71.0 & \cellcolor{tomato!80.0}80.0 & \cellcolor{tomato!68.0}68.0 & \cellcolor{tomato!29.7}29.7 & \cellcolor{tomato!6.2}6.2 & \cellcolor{tomato!98.8}98.8 & \cellcolor{tomato!9.3}9.3 & \cellcolor{tomato!15.2}15.2 & \cellcolor{tomato!66.1}66.1 & \cellcolor{tomato!90.0}90.0 & \cellcolor{tomato!15.3}15.3 & \cellcolor{tomato!78.6}78.6 & \cellcolor{tomato!19.9}19.9 & \cellcolor{tomato!44.3}44.3 \\
& 1290 & \cellcolor{tomato!59.1}59.1 & \cellcolor{tomato!66.0}66.0 & \cellcolor{tomato!78.0}78.0 & \cellcolor{tomato!70.0}70.0 & \cellcolor{tomato!18.6}18.6 & \cellcolor{tomato!8.4}8.4 & \cellcolor{tomato!94.2}94.2 & \cellcolor{tomato!7.5}7.5 & \cellcolor{tomato!10.3}10.3 & \cellcolor{tomato!69.0}69.0 & \cellcolor{tomato!87.0}87.0 & \cellcolor{tomato!14.4}14.4 & \cellcolor{tomato!77.1}77.1 & \cellcolor{tomato!23.5}23.5 & \cellcolor{tomato!35.4}35.4 \\
& 748 & \cellcolor{tomato!58.8}58.8 & \cellcolor{tomato!70.0}70.0 & \cellcolor{tomato!74.0}74.0 & \cellcolor{tomato!68.0}68.0 & \cellcolor{tomato!30.5}30.5 & \cellcolor{tomato!7.8}7.8 & \cellcolor{tomato!92.8}92.8 & \cellcolor{tomato!11.9}11.9 & \cellcolor{tomato!15.6}15.6 & \cellcolor{tomato!69.2}69.2 & \cellcolor{tomato!88.0}88.0 & \cellcolor{tomato!14.1}14.1 & \cellcolor{tomato!71.9}71.9 & \cellcolor{tomato!23.3}23.3 & \cellcolor{tomato!64.8}64.8 \\
\hline
\end{tabular}
\end{table*}

% ================== Qwen3-4B GainLoRA SuperNI ====================
\begin{table*}[t]
\centering
\caption{Performance dynamics during continual fine-tuning, using the \textbf{Qwen3-4B} model with the \textbf{GainLoRA} method (\textsc{Rouge} = 48.4, Forget Rate = 9.0).}
\renewcommand\arraystretch{0.97}
\setlength{\tabcolsep}{4pt}
\footnotesize
\label{gainloraseed1qwen3}
\begin{tabular}{|l|*{17}{c|}}
\hline
\multicolumn{2}{|c|}{} & \multicolumn{15}{c|}{Evaluation Task ID} \\
\cline{3-17}
\multicolumn{2}{|c|}{} & 591 & 073 & 1687 & 875 & 1572 & 639 & 1510 & 1590 & 511 & 181 & 363 & 1729 & 002 & 1290 & 748 \\
\hline
\multirow{15}{*}{\rotatebox{90}{Training Task ID}} 
& 591 & \cellcolor{tomato!67.8}67.8 & \cellcolor{tomato!79.0}79.0 & \cellcolor{tomato!78.0}78.0 & \cellcolor{tomato!43.0}43.0 & \cellcolor{tomato!34.3}34.3 & \cellcolor{tomato!4.0}4.0 & \cellcolor{tomato!95.7}95.7 & \cellcolor{tomato!12.8}12.8 & \cellcolor{tomato!14.6}14.6 & \cellcolor{tomato!12.3}12.3 & \cellcolor{tomato!88.0}88.0 & \cellcolor{tomato!11.9}11.9 & \cellcolor{tomato!77.1}77.1 & \cellcolor{tomato!19.5}19.5 & \cellcolor{tomato!43.6}43.6 \\
& 073 & \cellcolor{tomato!66.3}66.3 & \cellcolor{tomato!78.0}78.0 & \cellcolor{tomato!75.0}75.0 & \cellcolor{tomato!42.0}42.0 & \cellcolor{tomato!35.2}35.2 & \cellcolor{tomato!9.4}9.4 & \cellcolor{tomato!20.2}20.2 & \cellcolor{tomato!10.3}10.3 & \cellcolor{tomato!14.9}14.9 & \cellcolor{tomato!14.2}14.2 & \cellcolor{tomato!91.0}91.0 & \cellcolor{tomato!10.4}10.4 & \cellcolor{tomato!76.9}76.9 & \cellcolor{tomato!19.2}19.2 & \cellcolor{tomato!43.3}43.3 \\
& 1687 & \cellcolor{tomato!66.7}66.7 & \cellcolor{tomato!80.0}80.0 & \cellcolor{tomato!86.0}86.0 & \cellcolor{tomato!38.0}38.0 & \cellcolor{tomato!34.7}34.7 & \cellcolor{tomato!6.3}6.3 & \cellcolor{tomato!89.9}89.9 & \cellcolor{tomato!11.4}11.4 & \cellcolor{tomato!14.9}14.9 & \cellcolor{tomato!16.1}16.1 & \cellcolor{tomato!89.0}89.0 & \cellcolor{tomato!11.9}11.9 & \cellcolor{tomato!76.8}76.8 & \cellcolor{tomato!18.9}18.9 & \cellcolor{tomato!43.5}43.5 \\
& 875 & \cellcolor{tomato!61.1}61.1 & \cellcolor{tomato!66.0}66.0 & \cellcolor{tomato!82.0}82.0 & \cellcolor{tomato!83.0}83.0 & \cellcolor{tomato!37.2}37.2 & \cellcolor{tomato!5.3}5.3 & \cellcolor{tomato!92.7}92.7 & \cellcolor{tomato!14.0}14.0 & \cellcolor{tomato!13.9}13.9 & \cellcolor{tomato!6.2}6.2 & \cellcolor{tomato!79.0}79.0 & \cellcolor{tomato!11.4}11.4 & \cellcolor{tomato!75.2}75.2 & \cellcolor{tomato!18.3}18.3 & \cellcolor{tomato!42.6}42.6 \\
& 1572 & \cellcolor{tomato!55.2}55.2 & \cellcolor{tomato!69.0}69.0 & \cellcolor{tomato!80.0}80.0 & \cellcolor{tomato!77.0}77.0 & \cellcolor{tomato!44.8}44.8 & \cellcolor{tomato!4.3}4.3 & \cellcolor{tomato!93.3}93.3 & \cellcolor{tomato!9.1}9.1 & \cellcolor{tomato!14.5}14.5 & \cellcolor{tomato!28.2}28.2 & \cellcolor{tomato!82.0}82.0 & \cellcolor{tomato!15.9}15.9 & \cellcolor{tomato!76.5}76.5 & \cellcolor{tomato!18.7}18.7 & \cellcolor{tomato!43.6}43.6 \\
& 639 & \cellcolor{tomato!52.6}52.6 & \cellcolor{tomato!63.0}63.0 & \cellcolor{tomato!77.0}77.0 & \cellcolor{tomato!74.0}74.0 & \cellcolor{tomato!36.0}36.0 & \cellcolor{tomato!10.5}10.5 & \cellcolor{tomato!84.1}84.1 & \cellcolor{tomato!14.2}14.2 & \cellcolor{tomato!13.4}13.4 & \cellcolor{tomato!36.1}36.1 & \cellcolor{tomato!80.0}80.0 & \cellcolor{tomato!13.6}13.6 & \cellcolor{tomato!77.3}77.3 & \cellcolor{tomato!17.0}17.0 & \cellcolor{tomato!43.2}43.2 \\
& 1510 & \cellcolor{tomato!54.9}54.9 & \cellcolor{tomato!65.0}65.0 & \cellcolor{tomato!75.0}75.0 & \cellcolor{tomato!76.0}76.0 & \cellcolor{tomato!39.7}39.7 & \cellcolor{tomato!12.3}12.3 & \cellcolor{tomato!98.2}98.2 & \cellcolor{tomato!13.4}13.4 & \cellcolor{tomato!14.2}14.2 & \cellcolor{tomato!26.6}26.6 & \cellcolor{tomato!83.0}83.0 & \cellcolor{tomato!13.9}13.9 & \cellcolor{tomato!73.1}73.1 & \cellcolor{tomato!17.5}17.5 & \cellcolor{tomato!43.9}43.9 \\
& 1590 & \cellcolor{tomato!51.6}51.6 & \cellcolor{tomato!66.0}66.0 & \cellcolor{tomato!76.0}76.0 & \cellcolor{tomato!77.0}77.0 & \cellcolor{tomato!31.5}31.5 & \cellcolor{tomato!9.6}9.6 & \cellcolor{tomato!99.0}99.0 & \cellcolor{tomato!12.0}12.0 & \cellcolor{tomato!12.6}12.6 & \cellcolor{tomato!27.8}27.8 & \cellcolor{tomato!85.0}85.0 & \cellcolor{tomato!10.9}10.9 & \cellcolor{tomato!72.6}72.6 & \cellcolor{tomato!16.9}16.9 & \cellcolor{tomato!43.0}43.0 \\
& 511 & \cellcolor{tomato!50.9}50.9 & \cellcolor{tomato!66.0}66.0 & \cellcolor{tomato!81.0}81.0 & \cellcolor{tomato!72.0}72.0 & \cellcolor{tomato!28.3}28.3 & \cellcolor{tomato!4.8}4.8 & \cellcolor{tomato!98.3}98.3 & \cellcolor{tomato!7.0}7.0 & \cellcolor{tomato!14.9}14.9 & \cellcolor{tomato!29.7}29.7 & \cellcolor{tomato!84.0}84.0 & \cellcolor{tomato!8.7}8.7 & \cellcolor{tomato!73.5}73.5 & \cellcolor{tomato!20.2}20.2 & \cellcolor{tomato!37.5}37.5 \\
& 181 & \cellcolor{tomato!52.8}52.8 & \cellcolor{tomato!64.0}64.0 & \cellcolor{tomato!77.0}77.0 & \cellcolor{tomato!73.0}73.0 & \cellcolor{tomato!30.9}30.9 & \cellcolor{tomato!5.5}5.5 & \cellcolor{tomato!99.0}99.0 & \cellcolor{tomato!5.6}5.6 & \cellcolor{tomato!13.9}13.9 & \cellcolor{tomato!75.3}75.3 & \cellcolor{tomato!80.0}80.0 & \cellcolor{tomato!9.7}9.7 & \cellcolor{tomato!76.2}76.2 & \cellcolor{tomato!19.4}19.4 & \cellcolor{tomato!35.9}35.9 \\
& 363 & \cellcolor{tomato!50.8}50.8 & \cellcolor{tomato!65.0}65.0 & \cellcolor{tomato!74.0}74.0 & \cellcolor{tomato!70.0}70.0 & \cellcolor{tomato!21.0}21.0 & \cellcolor{tomato!4.3}4.3 & \cellcolor{tomato!99.7}99.7 & \cellcolor{tomato!5.5}5.5 & \cellcolor{tomato!13.8}13.8 & \cellcolor{tomato!64.1}64.1 & \cellcolor{tomato!91.0}91.0 & \cellcolor{tomato!11.1}11.1 & \cellcolor{tomato!76.0}76.0 & \cellcolor{tomato!18.9}18.9 & \cellcolor{tomato!36.7}36.7 \\
& 1729 & \cellcolor{tomato!51.1}51.1 & \cellcolor{tomato!66.0}66.0 & \cellcolor{tomato!79.0}79.0 & \cellcolor{tomato!64.0}64.0 & \cellcolor{tomato!27.8}27.8 & \cellcolor{tomato!6.5}6.5 & \cellcolor{tomato!99.0}99.0 & \cellcolor{tomato!8.0}8.0 & \cellcolor{tomato!12.4}12.4 & \cellcolor{tomato!67.3}67.3 & \cellcolor{tomato!89.0}89.0 & \cellcolor{tomato!13.7}13.7 & \cellcolor{tomato!77.9}77.9 & \cellcolor{tomato!18.7}18.7 & \cellcolor{tomato!25.2}25.2 \\
& 002 & \cellcolor{tomato!43.4}43.4 & \cellcolor{tomato!71.0}71.0 & \cellcolor{tomato!78.0}78.0 & \cellcolor{tomato!62.0}62.0 & \cellcolor{tomato!29.4}29.4 & \cellcolor{tomato!7.9}7.9 & \cellcolor{tomato!99.0}99.0 & \cellcolor{tomato!8.9}8.9 & \cellcolor{tomato!13.6}13.6 & \cellcolor{tomato!61.4}61.4 & \cellcolor{tomato!90.0}90.0 & \cellcolor{tomato!14.7}14.7 & \cellcolor{tomato!78.8}78.8 & \cellcolor{tomato!18.4}18.4 & \cellcolor{tomato!28.0}28.0 \\
& 1290 & \cellcolor{tomato!50.3}50.3 & \cellcolor{tomato!69.0}69.0 & \cellcolor{tomato!75.0}75.0 & \cellcolor{tomato!60.0}60.0 & \cellcolor{tomato!15.0}15.0 & \cellcolor{tomato!6.7}6.7 & \cellcolor{tomato!92.2}92.2 & \cellcolor{tomato!9.5}9.5 & \cellcolor{tomato!8.3}8.3 & \cellcolor{tomato!67.9}67.9 & \cellcolor{tomato!90.0}90.0 & \cellcolor{tomato!13.2}13.2 & \cellcolor{tomato!76.1}76.1 & \cellcolor{tomato!24.6}24.6 & \cellcolor{tomato!22.3}22.3 \\
& 748 & \cellcolor{tomato!48.1}48.1 & \cellcolor{tomato!67.0}67.0 & \cellcolor{tomato!76.0}76.0 & \cellcolor{tomato!54.0}54.0 & \cellcolor{tomato!25.9}25.9 & \cellcolor{tomato!7.5}7.5 & \cellcolor{tomato!99.0}99.0 & \cellcolor{tomato!11.9}11.9 & \cellcolor{tomato!10.8}10.8 & \cellcolor{tomato!62.4}62.4 & \cellcolor{tomato!89.0}89.0 & \cellcolor{tomato!13.9}13.9 & \cellcolor{tomato!71.4}71.4 & \cellcolor{tomato!24.3}24.3 & \cellcolor{tomato!65.3}65.3 \\
\hline
\end{tabular}
\end{table*}

\vspace{-3pt}
\section{Limitations}
\label{sec:limitations}
\vspace{-3pt}
While \baby effectively addresses the catastrophic forgetting in continual LLM fine-tuning without requiring access to original pre-training data, it possesses a few limitations that warrant future exploration. 
First, the differentiable generation mechanism naturally introduces additional computational overhead during the training phase. Although \baby completely circumvents the massive storage burden of maintaining historical pre-training datasets, auto-regressively generating pseudo data via Gumbel-Softmax increases the latency of the forward and backward passes compared to standard LoRA methods. We provide a comprehensive computational complexity analysis in Sec.~\ref{sec:complexity_analysis} and introduce several implementation refinements to minimize the overhead in Sec.~\ref{app:implementation}. 

Second, the representational capacity of the generated pseudo data sequence is inherently bounded by the predefined length of the soft prompts $L$ and generation steps $M$, and we provide a sensitivity experiment in Sec.~\ref{sec:sensitivity}. While our adversarial virtual step approximation guarantees that these sequences precisely target the most vulnerable boundaries of general-purpose knowledge, dealing with extremely long and highly heterogeneous task sequences might eventually saturate this representational capacity. Future work could explore dynamically expanding prompt lengths or incorporating hierarchical prompt routing to further scale the generation mechanism.

\vspace{-3pt}
\section{LLM Usage} \label{sec:llmusage}
\vspace{-3pt}
LLMs were utilized solely for grammatical refinement and linguistic polishing of the manuscript; the core ideas, methodology, and results were developed independently by the authors.

\clearpage
\section*{NeurIPS Paper Checklist}

\begin{enumerate}

\item {\bf Claims}
    \item[] Question: Do the main claims made in the abstract and introduction accurately reflect the paper's contributions and scope?
    \item[] Answer: \answerYes{} % Replace by \answerYes{}, \answerNo{}, or \answerNA{}.
    \item[] Justification: \textit{Our primary motivation is the observation that orthogonal gradient projection methods fail to preserve general-purpose knowledge due to unknown pre-training gradients. Therefore, we introduce \baby to estimate and orthogonalize these unknown pre-training gradients. These contributions are detailed extensively in our Abstract and Introduction.}
    \item[] Guidelines:
    \begin{itemize}
        \item The answer \answerNA{} means that the abstract and introduction do not include the claims made in the paper.
        \item The abstract and/or introduction should clearly state the claims made, including the contributions made in the paper and important assumptions and limitations. A \answerNo{} or \answerNA{} answer to this question will not be perceived well by the reviewers. 
        \item The claims made should match theoretical and experimental results, and reflect how much the results can be expected to generalize to other settings. 
        \item It is fine to include aspirational goals as motivation as long as it is clear that these goals are not attained by the paper. 
    \end{itemize}

\item {\bf Limitations}
    \item[] Question: Does the paper discuss the limitations of the work performed by the authors?
    \item[] Answer: \answerYes{} % Replace by \answerYes{}, \answerNo{}, or \answerNA{}.
    \item[] Justification: \textit{In Appendix~\ref{sec:limitations}, we provide a thorough discussion regarding the potential limitations of this work, focusing primarily on computational efficiency, which remains a critical area for prospective investigation.}
    \item[] Guidelines:
    \begin{itemize}
        \item The answer \answerNA{} means that the paper has no limitation while the answer \answerNo{} means that the paper has limitations, but those are not discussed in the paper. 
        \item The authors are encouraged to create a separate ``Limitations'' section in their paper.
        \item The paper should point out any strong assumptions and how robust the results are to violations of these assumptions (e.g., independence assumptions, noiseless settings, model well-specification, asymptotic approximations only holding locally). The authors should reflect on how these assumptions might be violated in practice and what the implications would be.
        \item The authors should reflect on the scope of the claims made, e.g., if the approach was only tested on a few datasets or with a few runs. In general, empirical results often depend on implicit assumptions, which should be articulated.
        \item The authors should reflect on the factors that influence the performance of the approach. For example, a facial recognition algorithm may perform poorly when image resolution is low or images are taken in low lighting. Or a speech-to-text system might not be used reliably to provide closed captions for online lectures because it fails to handle technical jargon.
        \item The authors should discuss the computational efficiency of the proposed algorithms and how they scale with dataset size.
        \item If applicable, the authors should discuss possible limitations of their approach to address problems of privacy and fairness.
        \item While the authors might fear that complete honesty about limitations might be used by reviewers as grounds for rejection, a worse outcome might be that reviewers discover limitations that aren't acknowledged in the paper. The authors should use their best judgment and recognize that individual actions in favor of transparency play an important role in developing norms that preserve the integrity of the community. Reviewers will be specifically instructed to not penalize honesty concerning limitations.
    \end{itemize}

\item {\bf Theory assumptions and proofs}
    \item[] Question: For each theoretical result, does the paper provide the full set of assumptions and a complete (and correct) proof?
    \item[] Answer: \answerYes{} % Replace by \answerYes{}, \answerNo{}, or \answerNA{}.
    \item[] Justification: \textit{In Appendix~\ref{sec:appendix_proofs}, we provide rigorous theoretical proofs for the claims and underlying assumptions, specifically addressing the three primary theoretical questions, presented throughout this work.}
    \item[] Guidelines:
    \begin{itemize}
        \item The answer \answerNA{} means that the paper does not include theoretical results. 
        \item All the theorems, formulas, and proofs in the paper should be numbered and cross-referenced.
        \item All assumptions should be clearly stated or referenced in the statement of any theorems.
        \item The proofs can either appear in the main paper or the supplemental material, but if they appear in the supplemental material, the authors are encouraged to provide a short proof sketch to provide intuition. 
        \item Inversely, any informal proof provided in the core of the paper should be complemented by formal proofs provided in appendix or supplemental material.
        \item Theorems and Lemmas that the proof relies upon should be properly referenced. 
    \end{itemize}

    \item {\bf Experimental result reproducibility}
    \item[] Question: Does the paper fully disclose all the information needed to reproduce the main experimental results of the paper to the extent that it affects the main claims and/or conclusions of the paper (regardless of whether the code and data are provided or not)?
    \item[] Answer: \answerYes{} % Replace by \answerYes{}, \answerNo{}, or \answerNA{}.
    \item[] Justification: \textit{In Sec.~\ref{sec:experiments}, we present comprehensive experimental results and analyses. Furthermore, Appendix~\ref{sec:appendix.setup} is dedicated to detailing the essential information of our experiments, including LLM architectures, baseline methods, datasets, and implementation details.}
    \item[] Guidelines:
    \begin{itemize}
        \item The answer \answerNA{} means that the paper does not include experiments.
        \item If the paper includes experiments, a \answerNo{} answer to this question will not be perceived well by the reviewers: Making the paper reproducible is important, regardless of whether the code and data are provided or not.
        \item If the contribution is a dataset and\slash or model, the authors should describe the steps taken to make their results reproducible or verifiable. 
        \item Depending on the contribution, reproducibility can be accomplished in various ways. For example, if the contribution is a novel architecture, describing the architecture fully might suffice, or if the contribution is a specific model and empirical evaluation, it may be necessary to either make it possible for others to replicate the model with the same dataset, or provide access to the model. In general. releasing code and data is often one good way to accomplish this, but reproducibility can also be provided via detailed instructions for how to replicate the results, access to a hosted model (e.g., in the case of a large language model), releasing of a model checkpoint, or other means that are appropriate to the research performed.
        \item While NeurIPS does not require releasing code, the conference does require all submissions to provide some reasonable avenue for reproducibility, which may depend on the nature of the contribution. For example
        \begin{enumerate}
            \item If the contribution is primarily a new algorithm, the paper should make it clear how to reproduce that algorithm.
            \item If the contribution is primarily a new model architecture, the paper should describe the architecture clearly and fully.
            \item If the contribution is a new model (e.g., a large language model), then there should either be a way to access this model for reproducing the results or a way to reproduce the model (e.g., with an open-source dataset or instructions for how to construct the dataset).
            \item We recognize that reproducibility may be tricky in some cases, in which case authors are welcome to describe the particular way they provide for reproducibility. In the case of closed-source models, it may be that access to the model is limited in some way (e.g., to registered users), but it should be possible for other researchers to have some path to reproducing or verifying the results.
        \end{enumerate}
    \end{itemize}

\item {\bf Open access to data and code}
    \item[] Question: Does the paper provide open access to the data and code, with sufficient instructions to faithfully reproduce the main experimental results, as described in supplemental material?
    \item[] Answer: \answerYes{} % Replace by \answerYes{}, \answerNo{}, or \answerNA{}.
    \item[] Justification: \textit{We have provided the source code for our implementation, accessible via the link included in Sec.~\ref{sec:1}.}
    \item[] Guidelines:
    \begin{itemize}
        \item The answer \answerNA{} means that paper does not include experiments requiring code.
        \item Please see the NeurIPS code and data submission guidelines (\url{https://neurips.cc/public/guides/CodeSubmissionPolicy}) for more details.
        \item While we encourage the release of code and data, we understand that this might not be possible, so \answerNo{} is an acceptable answer. Papers cannot be rejected simply for not including code, unless this is central to the contribution (e.g., for a new open-source benchmark).
        \item The instructions should contain the exact command and environment needed to run to reproduce the results. See the NeurIPS code and data submission guidelines (\url{https://neurips.cc/public/guides/CodeSubmissionPolicy}) for more details.
        \item The authors should provide instructions on data access and preparation, including how to access the raw data, preprocessed data, intermediate data, and generated data, etc.
        \item The authors should provide scripts to reproduce all experimental results for the new proposed method and baselines. If only a subset of experiments are reproducible, they should state which ones are omitted from the script and why.
        \item At submission time, to preserve anonymity, the authors should release anonymized versions (if applicable).
        \item Providing as much information as possible in supplemental material (appended to the paper) is recommended, but including URLs to data and code is permitted.
    \end{itemize}

\item {\bf Experimental setting/details}
    \item[] Question: Does the paper specify all the training and test details (e.g., data splits, hyperparameters, how they were chosen, type of optimizer) necessary to understand the results?
    \item[] Answer: \answerYes{} % Replace by \answerYes{}, \answerNo{}, or \answerNA{}.
    \item[] Justification: \textit{Comprehensive details regarding the experimental setup are provided in Appendix~\ref{sec:appendix.setup}. Furthermore, we perform a sensitivity analysis in Sec.~\ref{sec:sensitivity} to delineate the selection process for key hyper-parameters, \eg $L$ and $M$.}
    \item[] Guidelines:
    \begin{itemize}
        \item The answer \answerNA{} means that the paper does not include experiments.
        \item The experimental setting should be presented in the core of the paper to a level of detail that is necessary to appreciate the results and make sense of them.
        \item The full details can be provided either with the code, in appendix, or as supplemental material.
    \end{itemize}

\item {\bf Experiment statistical significance}
    \item[] Question: Does the paper report error bars suitably and correctly defined or other appropriate information about the statistical significance of the experiments?
    \item[] Answer: \answerYes{} % Replace by \answerYes{}, \answerNo{}, or \answerNA{}.
    \item[] Justification: \textit{The experiments presented in Table~\ref{result} were conducted across three independent trials using distinct random seeds, each corresponding to a different task order. We report the aggregated results from these trials, which consistently demonstrate statistical significance.}
    \item[] Guidelines:
    \begin{itemize}
        \item The answer \answerNA{} means that the paper does not include experiments.
        \item The authors should answer \answerYes{} if the results are accompanied by error bars, confidence intervals, or statistical significance tests, at least for the experiments that support the main claims of the paper.
        \item The factors of variability that the error bars are capturing should be clearly stated (for example, train/test split, initialization, random drawing of some parameter, or overall run with given experimental conditions).
        \item The method for calculating the error bars should be explained (closed form formula, call to a library function, bootstrap, etc.)
        \item The assumptions made should be given (e.g., Normally distributed errors).
        \item It should be clear whether the error bar is the standard deviation or the standard error of the mean.
        \item It is OK to report 1-sigma error bars, but one should state it. The authors should preferably report a 2-sigma error bar than state that they have a 96\% CI, if the hypothesis of Normality of errors is not verified.
        \item For asymmetric distributions, the authors should be careful not to show in tables or figures symmetric error bars that would yield results that are out of range (e.g., negative error rates).
        \item If error bars are reported in tables or plots, the authors should explain in the text how they were calculated and reference the corresponding figures or tables in the text.
    \end{itemize}

\item {\bf Experiments compute resources}
    \item[] Question: For each experiment, does the paper provide sufficient information on the computer resources (type of compute workers, memory, time of execution) needed to reproduce the experiments?
    \item[] Answer: \answerYes{} % Replace by \answerYes{}, \answerNo{}, or \answerNA{}.
    \item[] Justification: \textit{Appendix~\ref{subsec:empirical_budget} provides a comprehensive breakdown of the computational runtime measured across our 8-GPU cluster.}
    \item[] Guidelines:
    \begin{itemize}
        \item The answer \answerNA{} means that the paper does not include experiments.
        \item The paper should indicate the type of compute workers CPU or GPU, internal cluster, or cloud provider, including relevant memory and storage.
        \item The paper should provide the amount of compute required for each of the individual experimental runs as well as estimate the total compute. 
        \item The paper should disclose whether the full research project required more compute than the experiments reported in the paper (e.g., preliminary or failed experiments that didn't make it into the paper). 
    \end{itemize}
    
\item {\bf Code of ethics}
    \item[] Question: Does the research conducted in the paper conform, in every respect, with the NeurIPS Code of Ethics \url{https://neurips.cc/public/EthicsGuidelines}?
    \item[] Answer: \answerYes{} % Replace by \answerYes{}, \answerNo{}, or \answerNA{}.
    \item[] Justification: \textit{We have thoroughly reviewed the Code of Ethics and certify that our research adheres to all prescribed ethical standards and institutional guidelines.}
    \item[] Guidelines:
    \begin{itemize}
        \item The answer \answerNA{} means that the authors have not reviewed the NeurIPS Code of Ethics.
        \item If the authors answer \answerNo, they should explain the special circumstances that require a deviation from the Code of Ethics.
        \item The authors should make sure to preserve anonymity (e.g., if there is a special consideration due to laws or regulations in their jurisdiction).
    \end{itemize}

\item {\bf Broader impacts}
    \item[] Question: Does the paper discuss both potential positive societal impacts and negative societal impacts of the work performed?
    \item[] Answer: \answerNA{} % Replace by \answerYes{}, \answerNo{}, or \answerNA{}.
    \item[] Justification: \textit{This research focuses on the continual LLM fine-tuning, utilizing exclusively open-source instruction-tuning datasets, such as those for sentiment classification and text summarization. Consequently, this work does not present direct negative societal impacts regarding data privacy or sensitive information.}
    \item[] Guidelines:
    \begin{itemize}
        \item The answer \answerNA{} means that there is no societal impact of the work performed.
        \item If the authors answer \answerNA{} or \answerNo, they should explain why their work has no societal impact or why the paper does not address societal impact.
        \item Examples of negative societal impacts include potential malicious or unintended uses (e.g., disinformation, generating fake profiles, surveillance), fairness considerations (e.g., deployment of technologies that could make decisions that unfairly impact specific groups), privacy considerations, and security considerations.
        \item The conference expects that many papers will be foundational research and not tied to particular applications, let alone deployments. However, if there is a direct path to any negative applications, the authors should point it out. For example, it is legitimate to point out that an improvement in the quality of generative models could be used to generate Deepfakes for disinformation. On the other hand, it is not needed to point out that a generic algorithm for optimizing neural networks could enable people to train models that generate Deepfakes faster.
        \item The authors should consider possible harms that could arise when the technology is being used as intended and functioning correctly, harms that could arise when the technology is being used as intended but gives incorrect results, and harms following from (intentional or unintentional) misuse of the technology.
        \item If there are negative societal impacts, the authors could also discuss possible mitigation strategies (e.g., gated release of models, providing defenses in addition to attacks, mechanisms for monitoring misuse, mechanisms to monitor how a system learns from feedback over time, improving the efficiency and accessibility of ML).
    \end{itemize}
    
\item {\bf Safeguards}
    \item[] Question: Does the paper describe safeguards that have been put in place for responsible release of data or models that have a high risk for misuse (e.g., pre-trained language models, image generators, or scraped datasets)?
    \item[] Answer: \answerNA{} % Replace by \answerYes{}, \answerNo{}, or \answerNA{}.
    \item[] Justification: \textit{Our experiments are conducted using exclusively publicly available datasets and open-source LLMs. Consequently, this work presents no foreseeable risk regarding the misuse of models or datasets, as all resources are utilized in accordance with their respective open-source licenses and intended research purposes.}
    \item[] Guidelines:
    \begin{itemize}
        \item The answer \answerNA{} means that the paper poses no such risks.
        \item Released models that have a high risk for misuse or dual-use should be released with necessary safeguards to allow for controlled use of the model, for example by requiring that users adhere to usage guidelines or restrictions to access the model or implementing safety filters. 
        \item Datasets that have been scraped from the Internet could pose safety risks. The authors should describe how they avoided releasing unsafe images.
        \item We recognize that providing effective safeguards is challenging, and many papers do not require this, but we encourage authors to take this into account and make a best faith effort.
    \end{itemize}

\item {\bf Licenses for existing assets}
    \item[] Question: Are the creators or original owners of assets (e.g., code, data, models), used in the paper, properly credited and are the license and terms of use explicitly mentioned and properly respected?
    \item[] Answer: \answerYes{} % Replace by \answerYes{}, \answerNo{}, or \answerNA{}.
    \item[] Justification: \textit{All datasets and LLMs employed in this study are open-source, and their respective origins have been duly cited. Consequently, there are no unresolved licensing issues, as our use of these resources strictly adheres to their original distribution terms.}
    \item[] Guidelines:
    \begin{itemize}
        \item The answer \answerNA{} means that the paper does not use existing assets.
        \item The authors should cite the original paper that produced the code package or dataset.
        \item The authors should state which version of the asset is used and, if possible, include a URL.
        \item The name of the license (e.g., CC-BY 4.0) should be included for each asset.
        \item For scraped data from a particular source (e.g., website), the copyright and terms of service of that source should be provided.
        \item If assets are released, the license, copyright information, and terms of use in the package should be provided. For popular datasets, \url{paperswithcode.com/datasets} has curated licenses for some datasets. Their licensing guide can help determine the license of a dataset.
        \item For existing datasets that are re-packaged, both the original license and the license of the derived asset (if it has changed) should be provided.
        \item If this information is not available online, the authors are encouraged to reach out to the asset's creators.
    \end{itemize}

\item {\bf New assets}
    \item[] Question: Are new assets introduced in the paper well documented and is the documentation provided alongside the assets?
    \item[] Answer: \answerYes{} % Replace by \answerYes{}, \answerNo{}, or \answerNA{}.
    \item[] Justification: \textit{We have open-sourced the implementation code for our experiments and provided comprehensive documentation regarding its usage.}
    \item[] Guidelines:
    \begin{itemize}
        \item The answer \answerNA{} means that the paper does not release new assets.
        \item Researchers should communicate the details of the dataset\slash code\slash model as part of their submissions via structured templates. This includes details about training, license, limitations, etc. 
        \item The paper should discuss whether and how consent was obtained from people whose asset is used.
        \item At submission time, remember to anonymize your assets (if applicable). You can either create an anonymized URL or include an anonymized zip file.
    \end{itemize}

\item {\bf Crowdsourcing and research with human subjects}
    \item[] Question: For crowdsourcing experiments and research with human subjects, does the paper include the full text of instructions given to participants and screenshots, if applicable, as well as details about compensation (if any)? 
    \item[] Answer: \answerNA{} % Replace by \answerYes{}, \answerNo{}, or \answerNA{}.
    \item[] Justification: \textit{Our experiments do not require any additional crowdsourced data collection; all analyses are conducted using existing, publicly available benchmarks.}
    \item[] Guidelines:
    \begin{itemize}
        \item The answer \answerNA{} means that the paper does not involve crowdsourcing nor research with human subjects.
        \item Including this information in the supplemental material is fine, but if the main contribution of the paper involves human subjects, then as much detail as possible should be included in the main paper. 
        \item According to the NeurIPS Code of Ethics, workers involved in data collection, curation, or other labor should be paid at least the minimum wage in the country of the data collector. 
    \end{itemize}

\item {\bf Institutional review board (IRB) approvals or equivalent for research with human subjects}
    \item[] Question: Does the paper describe potential risks incurred by study participants, whether such risks were disclosed to the subjects, and whether Institutional Review Board (IRB) approvals (or an equivalent approval/review based on the requirements of your country or institution) were obtained?
    \item[] Answer: \answerNA{} % Replace by \answerYes{}, \answerNo{}, or \answerNA{}.
    \item[] Justification: \textit{Our experiments do not require any additional crowdsourced data collection; all analyses are conducted using existing, publicly available benchmarks.}
    \item[] Guidelines:
    \begin{itemize}
        \item The answer \answerNA{} means that the paper does not involve crowdsourcing nor research with human subjects.
        \item Depending on the country in which research is conducted, IRB approval (or equivalent) may be required for any human subjects research. If you obtained IRB approval, you should clearly state this in the paper. 
        \item We recognize that the procedures for this may vary significantly between institutions and locations, and we expect authors to adhere to the NeurIPS Code of Ethics and the guidelines for their institution. 
        \item For initial submissions, do not include any information that would break anonymity (if applicable), such as the institution conducting the review.
    \end{itemize}

\item {\bf Declaration of LLM usage}
    \item[] Question: Does the paper describe the usage of LLMs if it is an important, original, or non-standard component of the core methods in this research? Note that if the LLM is used only for writing, editing, or formatting purposes and does \emph{not} impact the core methodology, scientific rigor, or originality of the research, declaration is not required.
    %this research? 
    \item[] Answer: \answerYes{} % Replace by \answerYes{}, \answerNo{}, or \answerNA{}.
    \item[] Justification: \textit{In Appendix~\ref{sec:llmusage}, we disclose that LLMs were employed strictly for grammatical proofreading and linguistic refinement to enhance the readability of the manuscript.}
    \item[] Guidelines:
    \begin{itemize}
        \item The answer \answerNA{} means that the core method development in this research does not involve LLMs as any important, original, or non-standard components.
        \item Please refer to our LLM policy in the NeurIPS handbook for what should or should not be described.
    \end{itemize}

\end{enumerate}

\end{document}